\documentclass[acmsmall]{acmart}

\AtBeginDocument{%
  }

\setcopyright{acmcopyright}
\copyrightyear{2018}
\acmYear{2018}
\acmDOI{XXXXXXX.XXXXXXX}

\acmJournal{JACM}
\acmVolume{37}
\acmNumber{4}
\acmArticle{111}
\acmMonth{8}

\usepackage{amsmath} 
\usepackage{color}
\usepackage{multirow}
\usepackage{subcaption}
\begin{document}

\title{DiRaC-I: Identifying Diverse and Rare Training Classes for Zero-Shot Learning}

\author{Sandipan Sarma}
\email{sandipan.sarma@iitg.ac.in}
\author{Arijit Sur}
\email{arijit@iitg.ac.in}
\affiliation{%
  \institution{Indian Institute of Technology Guwahati}
  \streetaddress{Multimedia Lab, Department of Computer Science and Engineering}
  \city{Guwahati}
  \state{Assam}
  \country{India}
  \postcode{781039}
}

%
%
%
%
%
%

\renewcommand{\shortauthors}{Sandipan Sarma and Arijit Sur}

\begin{abstract}
  Inspired by strategies like Active Learning, it is intuitive that intelligently selecting the training classes from a dataset for Zero-Shot Learning (ZSL) can improve the performance of existing ZSL methods. In this work, we propose a framework called Diverse and Rare Class Identifier (DiRaC-I) which, given an attribute-based dataset, can intelligently yield the most suitable ``seen classes'' for training ZSL models. DiRaC-I has two main goals -- constructing a diversified set of seed classes, followed by a visual-semantic mining algorithm initialized by these seed classes that acquires the classes capturing both diversity and rarity in the object domain adequately. These classes can then be used as “seen classes” to train ZSL models for image classification. We adopt a real-world scenario where novel object classes are available to neither DiRaC-I nor the ZSL models during training and conducted extensive experiments on two benchmark data sets for zero-shot image classification — CUB and SUN. Our results demonstrate DiRaC-I helps ZSL models to achieve significant classification accuracy improvements.
\end{abstract}

\ccsdesc[500]{Computing methodologies~Transfer learning}

\keywords{Zero-shot learning, deep learning, object recognition, image classification}

\maketitle

\section{Introduction}
\label{sec:intro}
\begin{figure}[t]
	\centering
	\begin{subfigure}[t]{0.35\textwidth}
		\includegraphics[scale=0.45]{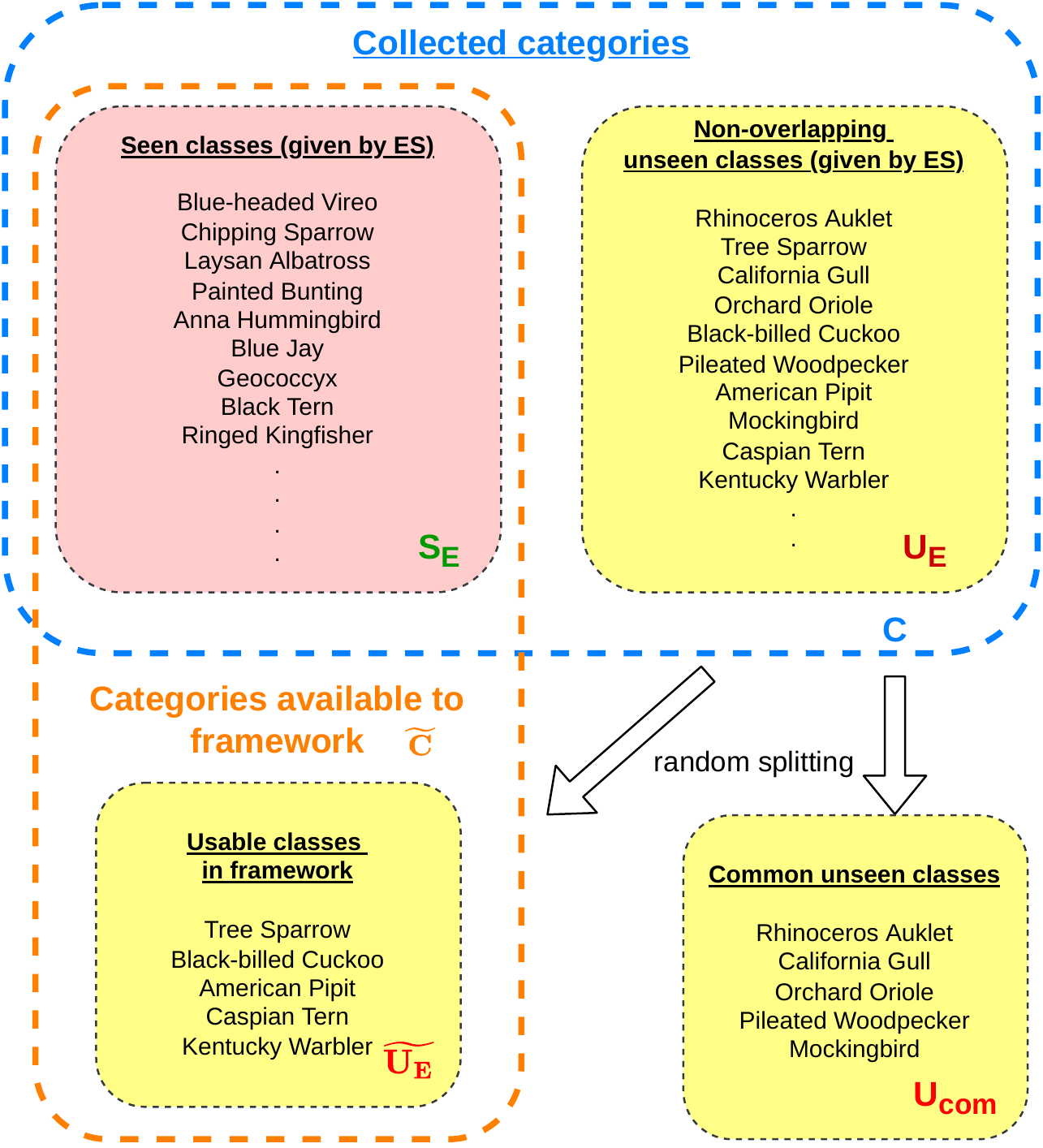}
		\caption{Splitting the available set of classes in a collected dataset to obtain an object domain to work with and the novel classes for model evaluation}
		\label{fig:test split}
	\end{subfigure} 
	\hfil
	\begin{subfigure}[t]{0.45\textwidth}
		\includegraphics[scale=0.45]{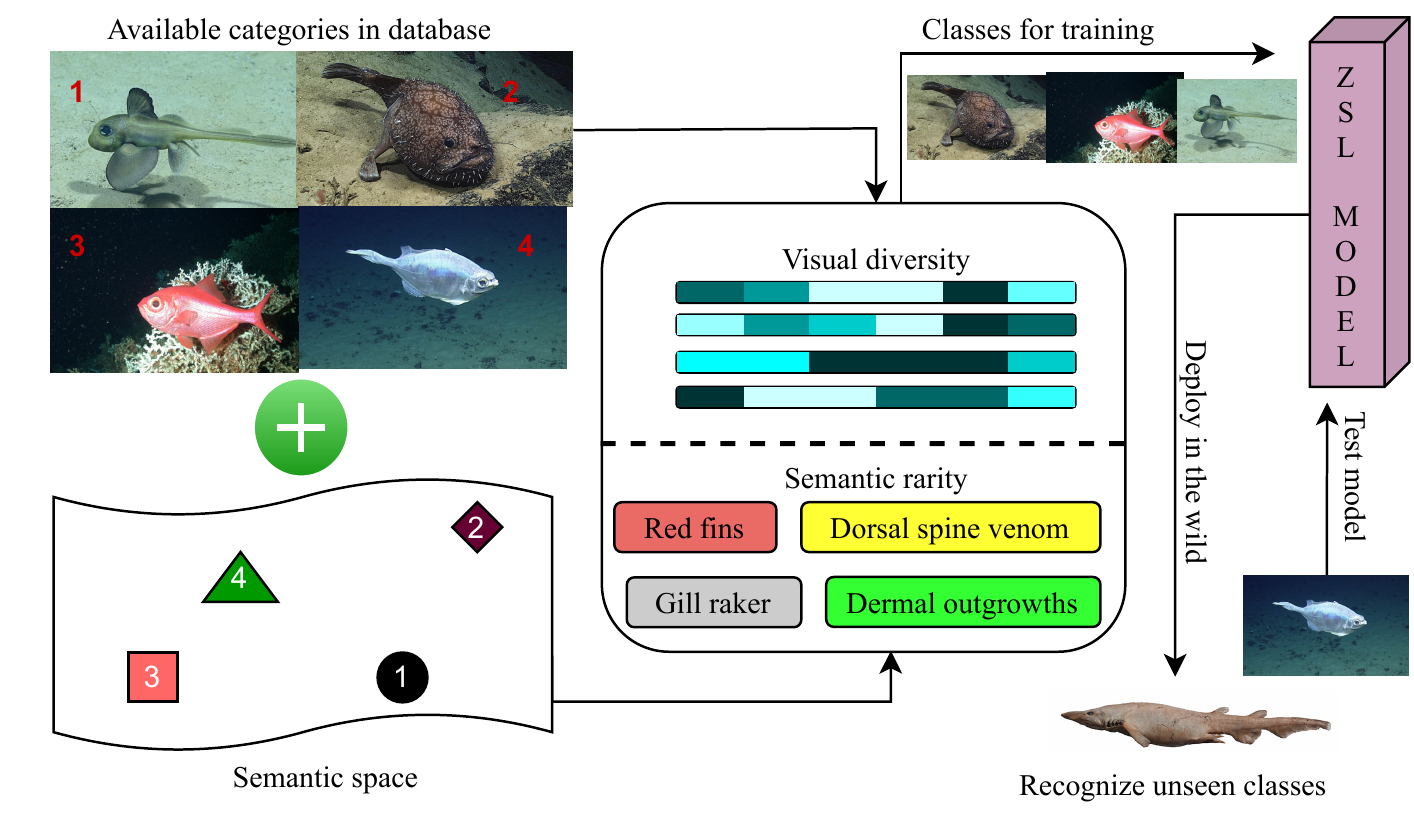}
		\caption{``Seen classes'' designated by DiRaC-I for training ZSL models -- an application}
		\label{fig:work_idea}
	\end{subfigure}
	\caption{\textbf{The target of DiRaC-I and its application}. \textit{(Left)} In a real-world scenario, data for a certain number of object categories are collected by dataset constructors. For example, Existing Split (ES) proposed by Xian et al.~\cite{xian2018zero} defines some fixed classes to be used as ``seen'' ($\mathcal{S}\textsubscript{E}$) and ``unseen'' ($\mathcal{U}\textsubscript{E}$) while performing ZSL. We randomly split the set $\mathcal{U}\textsubscript{E}$ and obtain a set of common unseen classes ($\mathcal{U}_{com}$). The proposed framework has access to only the remaining classes belonging to set $\Tilde{C}$, which constitute the known object domain for the framework. After training several existing ZSL models with seen classes both from ES and PS (our Proposed Splits), model performance is evaluated only on classes from $\mathcal{U}_{com}$ for fair comparison, because these classes are common to both ES and PS. We can consider classes from $\mathcal{U}_{com}$ to act as novel classes encountered {\it in the wild}; \textit{(Right)} Leveraging the DiRaC-I framework for selecting suitable seen classes from the object domain of {\it fish} -- a real-world application}
	\Description{How to split the classes from a dataset into disjoint sets and select a suitable set of seen classes for ZSL. An application of our framework DiRaC-I ZSL model while exploring underwater biodiversity is also shown, leveraging visual and semantic spaces of the object domain.}
	\label{fig:split-n-idea}
\end{figure}

Object recognition has witnessed a significant improvement in the recent past using deep learning methods~\cite{simonyan2014very,szegedy2015going,he2016deep,huang2017densely,howard2017mobilenets,liu2021medical} trained on large, annotated data sets in a supervised fashion. However, such methods fail when novel concepts are encountered. For example, an underwater robot exploring deep-sea biodiversity should trigger an alert if it encounters a novel or rare species --- like a \textit{Manocherian's Catshark} (Fig.~\ref{fig:work_idea}) --- but would probably fail as its recognition model is not trained on visual images of that species. A human, on the contrary, can recognize it if he/she has a visual perception about sharks and is given additional information that it looks like a small shark with some characteristic {\it attributes} --- a whitish, porcelain-colored body with a white spot on the tail tip. The idea of zero-shot learning (ZSL)~\cite{lampert2013attribute,xian2018zero} stems from this ability of humans to recognize unseen objects by learning a mapping function associating the visual samples from the seen classes with their {\it semantics} (or attributes). This function is then used to recognize both seen and unseen objects. 

\begin{figure}[t]
	\centering
	\includegraphics[width=\linewidth]{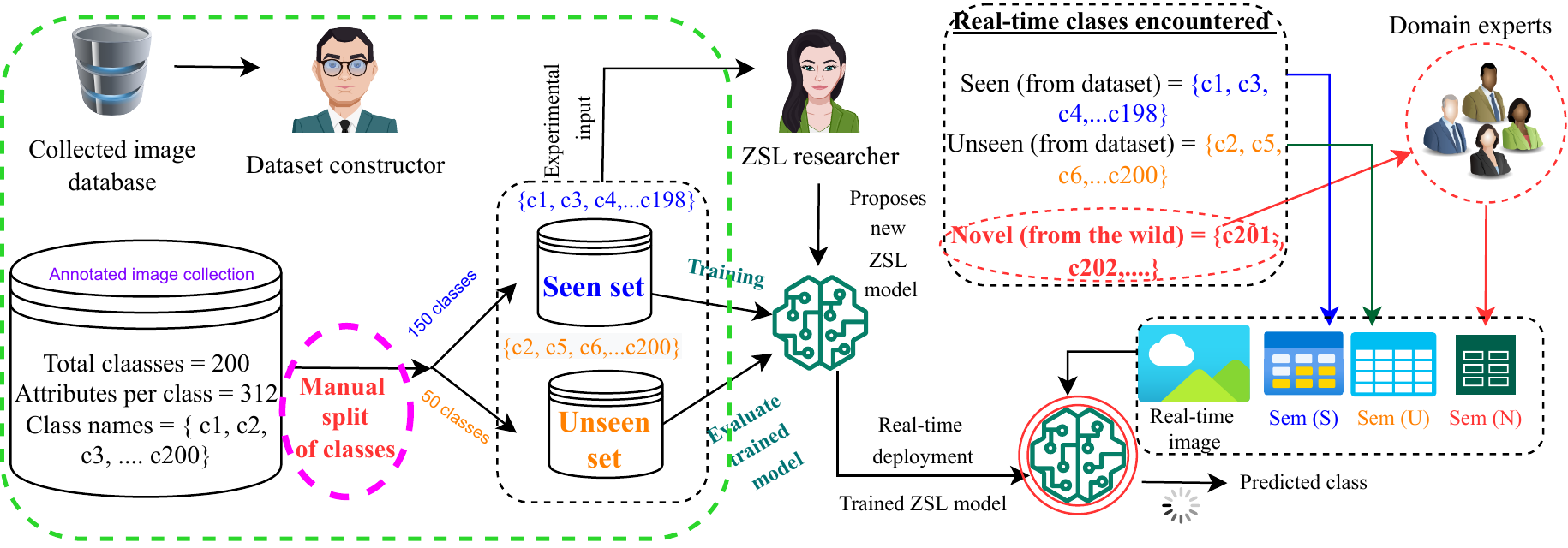}
	
	\caption{\textbf{ZSL community and the contributors.} Dataset constructors manually define split of seen-unseen classes which is received by ZSL researchers and used for training and evaluating ZSL models. High-performance models would be used for real-world deployment for image classification. {\bf DiRaC-I targets the work of the constructors ({\it green dotted box}) and aims to replace manual splits by intelligent splits automatically.}}
	\Description{Different actors of the ZSL community}
	\label{fig:zsl-comm}
\end{figure}

The ZSL community can be divided into three groups (Fig.~\ref{fig:zsl-comm}) based on their contributions towards ZSL -- (1) \textit{Dataset constructors}, who collect labeled data and semantics for a fixed number (say $k$) of object classes. Before releasing the dataset for ZSL research, they define a disjoint seen-unseen split of the $k$ classes {\bf manually}; (2) \textit{ZSL researchers}, who use these {\it predetermined} seen classes for training the ZSL models they propose, and the {\it predetermined} unseen classes to evaluate these models and simulate their ability to classify unseen classes {\it of the wild} when deployed in future; (3) \textit{AI-aided industries}, which deploy the state-of-the-art models to solve real-world problems using ZSL. For ZSL-based classification models to be widely accepted in the future by industries, the need for high-performance ZSL models trained with classes that capture rarity and diversity of the \textit{object domain} (defined in Sec.~\ref{dirac-i}) is paramount. However, ZSL models proposed by researchers today only {\it passively} learn from the predetermined set of seen classes provided by dataset constructors, like Xian et al.~\cite{xian2018zero}. They claim that class diversity is maintained while manually defining the seen-unseen sets; however, such splits are not designed for best zero-shot performance~\cite{lampert2013attribute}. Hence, an intelligent seen-unseen split of the $k$ classes of the collected dataset should be designed such that the designated seen classes automatically capture the diversity and rarity of the object domain. Only a few studies have addressed this issue in ZSL, mostly using Active Learning (AL) approaches. \cite{xie2016active,xie2017active} experiment with textual datasets only. Recently, in the image classification area,~\cite{wang2021graph} proposed an GCN-based AL framework for selecting the most crucial classes as seen classes for training. However, all these works initialize the AL algorithm with a randomly selected set of classes with labeled examples, called the {\it seed set}. Additionally, they do not consider the rare attributes for enriching the training set.

In this work, we propose a two-stage framework named \textit{Diverse and Rare Class Identifier (DiRaC-I)} inspired by AL which targets the attribute-based dataset constructors. From the $k$-class dataset provided by the constructors, DiRaC-I aims to select the most suitable seen classes for training ZSL models while trying to capture visual diversity and semantic rarity. The first stage is seed-set construction, where the $k$ classes are clustered based on semantic similarity. A single representative is picked from each cluster to ensure diversity while jointly prioritizing semantic rarity, forming a seed set. Intuitively, doing so would incorporate a generalized initial understanding of the object domain within the seed set, which is used as input for the next stage --- Visual-Semantic Mining (VSM). Here, we seek to maximize the diversity between visual samples of the seed and those of the other classes by estimating the distribution of {\it related} classes, based on the work by~\cite{bendale2016towards}, and select a few candidate non-seed classes. We define a {\it semantic score} to be computed for each of them, and a few classes having the highest semantic scores are added to the seed set. This process continues iteratively till we get a fixed number of seed classes, which become our seen classes to be provided as an output to researchers for training ZSL models.

We ensure a fair comparison of the knowledge gained while training using our seen classes (from Proposed seen-unseen Splits (PS)) and the predetermined (from Existing seen-unseen Splits (ES)) by evaluating the performance of existing ZSL models on a common set of unseen classes. For a given data set, this set is derived by randomly picking 50\% classes from all the unseen classes (for reasons given in Sec.~\ref{sec:3.2}) used in ES. These classes are not used during the entire operation of DiRaC-I or ZSL model training but only during the evaluation of ZSL models. Extensive experiments conducted on two challenging benchmark data sets --- CUB and SUN --- demonstrate that zero-shot accuracy of most models are enhanced when trained with seen classes acquired by DiRaC-I. In real-world situations, our framework should be able to select seen classes from a given attribute-based dataset to help improve the training of the ZSL models to be deployed for image classification. 

We summarize our contributions as follows. (1) We design a framework, DiRaC-I, that intelligently captures the diversity and rarity of the object domain within a set of seen classes on which ZSL models can be trained to have a comprehensive idea of the domain. (2) For initializing the VSM algorithm, a few diverse seed classes are selected as per some attribute-based scores, instead of just selecting them randomly. The rare attributes play a key role in computing these scores. (3) ZSL model performance is evaluated on a set of unseen classes common to both existing splits (ES) and proposed splits (PS) that are unavailable to DiRaC-I and the model during training (simulating practical scenarios of encountering novel classes in the wild). Hence, a fair comparison between the knowledge gained by the model trained with predetermined seen classes of ES and the acquired seen classes of PS is ensured. (4) Unlike~\cite{wang2021graph}, DiRaC-I can be used by ZSL researchers as a predecessor to select the seen classes from a given attribute-based dataset, and therefore can adapt to real-world scenarios. 

%

\section{Related Work}
\label{sec:literature}
\subsection{Zero-shot learning (ZSL)}
\label{sec:lit-zsl}
Motivated by the problems faced in supervised learning, several new learning paradigms have been proposed in the last decade or so, such as few-shot~\cite{cao2022learning,jiang2020few} and one-shot learning~\cite{fei2006one,liu2021single}. However, methods under these paradigms are still not able to cope with scenarios where we have ``zero" training samples of certain classes -- e.g. a rare fish species. Therefore, in the recent years there has been an increasing amount of interest in zero-shot learning, which defines a setting where visual features for unseen classes are unavailable during model training. However, seen and unseen classes can be linked through their semantics. ZSL has been applied to a wide array of computer vision tasks such as object detection~\cite{bansal2018zero}, action recognition~\cite{chen2021elaborative} and cross-modal retrieval~\cite{xu2021zero} to name a few. We, however, focus on the zero-shot image classification task in the following discussion.

The early works on ZSL~\cite{rohrbach2010helps,rohrbach2011evaluating,kankuekul2012online,lampert2013attribute} tried to learn intermediate attribute classifiers to transfer knowledge from the seen to unseen. Several other approaches that followed~\cite{frome2013devise,socher2013zero,akata2015label,akata2015evaluation,romera2015embarrassingly,xian2016latent,kodirov2017semantic} directly set up bi-linear and non-linear compatibility functions between the visual and semantic spaces. At test time, unseen visual features are projected to the semantic space using the learned functions, and the predicted class is the one achieving maximum compatibility score. The approach of learning a mixture model of seen classes to represent the images and semantic embeddings is taken up in~\cite{norouzi2013zero,zhang2015zero,changpinyo2016synthesized}. While these approaches work well in the conventional setting (CZSL) of unseen test classes only, in practice, a model should be able to classify samples from both seen and unseen classes when deployed. Generalized zero-shot learning (GZSL) is a setting that considers such a scenario. Most existing works that show improvements in GZSL have incorporated generative models~\cite{mishra2018generative,xian2018feature,felix2018multi,xian2019f,vyas2020leveraging,narayan2020latent,feng2020transfer,tang2021zero}, where the aim is to synthesize high-quality unseen class samples or visual features, converting the ZSL problem into a simple supervised classification task. A benchmark providing standard evaluation protocols and seen-unseen splits for some of the widely-used data sets in ZSL is given by~\cite{xian2018zero}. However, manually created seen-unseen splits might not capture the diversity and rarity well enough for training ZSL models, affecting their knowledge about the object domain.

\subsection{Selecting suitable seen classes for ZSL}
\label{sec:lit-al-zsl}
The idea of training ZSL models with seen classes more informative than the predetermined ones is relatively new. Recently, Active Learning (AL)~\cite{hanneke2014theory} strategies have been employed in this direction. However, contrary to the traditional way of acquiring the most informative {\it instances} from a data set for training classifiers, in the zero-shot setting, the objective of AL changes to acquiring informative {\it classes}. \cite{xie2016active} proposes a probabilistic method that focuses on two properties --- informativeness of the seen classes and their connectivity to the unseen. An extension of this work~\cite{xie2017active} demonstrates the impact of AL on ZSL for extreme multi-label classification. 
However, it experiments with textual data sets only. \cite{wang2021graph} adopts an AL approach for GCN-based zero-shot image classification. Their work extends the k-center algorithm with a Laplacian energy-based strategy for selecting the most crucial classes as seen classes. However, it is limited to GCN frameworks for ZSL and initializes the algorithm with a randomly selected seed set, like the other works on AL-based ZSL. 

Research in traditional AL has shown that instead of selecting the seed set {\it randomly}, an intelligent selection can propel AL in better directions. In an effort to justify this,~\cite{tomanek-etal-2009-proper} proposes to manually prepare the seed set, artificially enriched with rare class examples. \cite{dligach-palmer-2011-good} gives an automatic approach that follows the same principle. Nevertheless, to the best of our knowledge, no work has shown the combined benefits of intelligently acquiring seed classes and using them to obtain seen classes that capture the diversity and rarity from the object domain. Our proposed framework (DiRaC-I) first constructs a seed set with a diverse initial representation of the object domain. It then initializes an AL-inspired algorithm with this seed set and iteratively acquires a fixed number of seen classes using which ZSL models are to be trained. The most recent work with a similar objective picks the seed set randomly and is compatible only with GCN-based zero-shot frameworks~\cite{wang2021graph}. Moreover, their experiments are on a single dataset, and evaluation metrics are not comparable to the standard ones~\cite{xian2018zero}. On the other hand, DiRaC-I can work with any attribute-based data set in practical scenarios. We also evaluate the prediction accuracy of several existing ZSL models trained with seen classes from Existing and Proposed Splits using the standard metrics and obtain encouraging results.

\section{Problem setting and notations} 
\label{sec:ps-not}

\subsection{Object recognition using zero-shot learning}
\label{sec:3.1}

In a typical zero-shot setting, we have sets $\mathcal{S}$ and $\mathcal{U}$ of $N_{s}$ and $N_{u}$ number of seen and unseen classes respectively, such that $\mathcal{S} \, \cap \, \mathcal{U} = \emptyset$. Let $\mathcal{C} = \mathcal{S} \cup \mathcal{U}$ denote the set of all classes for a given data set. The associated semantic embeddings for these sets can be represented by $\mathcal{P}(\mathcal{S}) \in \mathbb{R}^{N_{s} \times d}$ and $\mathcal{P}(\mathcal{U}) \in \mathbb{R}^{N_{u} \times d}$ respectively, where attributes of a class $c$ are represented by a $d-$dimensional vector $\langle a_c^1, a_c^2,...a_c^d\rangle$. These embeddings are available in several forms like human-annotated attributes~\cite{xian2018zero}, word embeddings like Word2Vec~\cite{mikolov2013distributed} and GloVe~\cite{pennington-etal-2014-glove}, or hierarchical embeddings like WordNet~\cite{miller1995wordnet}. $\mathcal{X}^{s} \in \mathbb{R}^{m \times k}$ and $\mathcal{X}^{u} \in \mathbb{R}^{n \times k}$ represent the visual data for seen and unseen samples respectively, usually available in the form of visual features extracted from a CNN like ResNet-101~\cite{he2016deep,xian2018zero}, pretrained on a large-scale visual dataset like ImageNet~\cite{russakovsky2015imagenet}. $m$ and $n$ denote the number of seen and unseen class samples respectively, with each image being represented by a $k-$dimensional feature vector. Then, given training data $\mathcal{D} = \{(x_{j}^{s}, y_{j}^{s}) \in \mathcal{X}^{s} \times \mathcal{S} \}$ along with $\mathcal{P}(\mathcal{S})$ and $\mathcal{P}(\mathcal{U})$, the task in CZSL is to learn a classifier $f_{czsl} : \mathcal{X}^{u} \rightarrow \mathcal{U}$. In GZSL, a small subset of $\mathcal{X}^s$ ($\mathcal{X}^s_{sub}$) is used as the set of seen samples at test time. Then, the objective changes to learning a classifier $f_{gzsl} : \mathcal{X}^{s}_{sub} \cup \mathcal{X}^{u} \rightarrow \mathcal{S} \cup \mathcal{U}$ to classify both seen and unseen objects. 

\subsection{Practical insights into seen-unseen splits} 
\label{sec:3.2}
Scarcity of labeled data and dealing with unseen concepts are two potential areas where ZSL can contribute significantly in the future when deployed in practical applications. To name a few, with ZSL models: (1) autonomous vehicles~\cite{rezaei2014look,rajasekhar2015autonomous,ishihara2021multi} should be able to recognize unseen {\it concept cars} while driving; (2) previously unseen diseases like COVID-19 could be diagnosed based on their novel characteristics combined with the similarity to other known diseases like asthma~\cite{chen2021deep,rahman2021multimodal,rezaei2020zero}; (3) Autonomous Underwater Vehicles (AUVs) deployed in underwater explorations~\cite{kunz2008deep,kennedy2019unknown} should be able to recognize new fish or coral species if encountered (Fig.~\ref{fig:work_idea}).
However, for a target application, data for only a fixed number of available categories (comprising a set $\Tilde{\mathcal{C}}$) can be collected by the dataset constructors. Although they have access to labeled examples of all the $|\Tilde{\mathcal{C}}|$ classes and can provide them to ZSL researchers, the researchers cannot train their models with all $|\Tilde{\mathcal{C}}|$ classes as they would always need a disjoint set of unseen classes to evaluate ZSL models, as per ZSL criteria. Consequently, training ZSL models requires a subset $\mathcal{S} \subset \Tilde{\mathcal{C}}$ (i.e. the set of seen classes), which DiRaC-I helps the constructors to obtain. Classes from the other subset of the collected dataset($\mathcal{U} = \Tilde{\mathcal{C}} \setminus \mathcal{S}$) can be considered unseen classes by researchers to evaluate their model performance. Finally, the trained model can be deployed in the future to recognize novel classes (with known attributes) {\it in the wild} (Fig.~\ref{fig:work_idea}). 

For zero-shot classification, current researchers use seen-unseen splits predetermined by~\cite{xian2018zero}. However, unlike these {\bf Existing Splits (ES)}, we try to emulate the real-world scenario via our {\bf Proposed Splits (PS)}, where the seen classes exhibiting diversity and rarity can be automatically acquired from $\Tilde{\mathcal{C}}$ itself. For fair comparison of the knowledge gained by existing ZSL models when trained with seen classes from ES and PS, they should be evaluated on the same set of unseen classes. Since we do not have data from classes that are completely unknown to us during experimentation, we extract a few classes from the unseen set originally given by ES, and make them unavailable to both DiRaC-I and the ZSL models during their training. We first dissociate the set $\mathcal{U}$ of ES ($\mathcal{U}$\textsubscript{E}) into two halves randomly --- $\mathcal{U}_{com}$ becomes the set of $N_{u_{com}}$ unseen classes {\it of the wild} and $\Tilde{\mathcal{U}\textsubscript{E}}$ the other half, having $N_{\Tilde{u}}$ classes. Then, the proposed framework acquires seen classes from the set $\Tilde{\mathcal{C}} = \mathcal{S}\textsubscript{E} \, \cup \, \Tilde{\mathcal{U}\textsubscript{E}}$ and ZSL model can train on the acquired classes ($\mathcal{S}$\textsubscript{P}). Figure~\ref{fig:test split} gives a better understanding of this process. Finally, let $\Phi_E = \{ \Phi_{E}^{M_1}, \Phi_{E}^{M_2},... \Phi_{E}^{M_n}\}$ and $\Phi_P = \{ \Phi_{P}^{M_1}, \Phi_{P}^{M_2},...\Phi_{P}^{M_n} \}$ denote the sets of models $M_1,M_2,...M_n$ trained using seen classes from ES and PS respectively. We compare the performance of the models $\Phi_{E}^x$ and $\Phi_{P}^x$ on the test set $\mathcal{U}_{com}$ that is unseen to both $\Phi_{E}^x$ and $\Phi_{P}^x (x = M_1, M_2,...M_n)$. Note that in our framework, classes in $\mathcal{U}_{com}$ do not overlap with the ImageNet 1K classes used for pretraining ResNet-101, following the ZSL assumption provided by~\cite{xian2018zero}. Moreover, PS is not fixed -- since we induce randomness while splitting $\mathcal{U}$\textsubscript{E}, we repeat the entire process (from initializing DiRaC-I to evaluating ZSL models trained with the acquired seen classes) three times so that three different sets of classes are available to our framework at its inception. We show our results in each case, demonstrating the robustness of our framework to the available object domain. 


\section{DiRaC-I: Diverse and Rare Class Identifier}
\label{dirac-i}
In this work, we focus on data sets having homogeneous categories only --- e.g. having all bird categories. For such a data set, we say that its {\it object domain} is birds. Heuristically, training a ZSL model with seen classes that capture both the diversity in the visual space and rarity in the semantic space would provide it with a more generalized idea of the object domain. Hence, the key to our approach is exploring the entire available object domain (defined by classes from $\Tilde{\mathcal{C}}$) for diversity and rarity using a method {\it inspired by} Active Learning. Adopting such a principle enhances the capability of ZSL models for knowledge transfer from the seen to unseen classes during evaluation. Moreover, the novel classes exhibiting rare attributes have a better chance of being recognized accurately, as suggested by the results of our experiments on two benchmark data sets (Tab.~\ref{tab:2}). DiRaC-I consists of two stages, which are discussed in the following sections.

\subsection{Stage 1: Seed-set construction}
\label{sec:seed}

Let $\Tilde{\Psi}_i = \{ \Psi_{i}^1, \Psi_{i}^2...\Psi_{i}^i\}$ denote the set of $i$ clusters, where $\Psi_{i}^j$ denotes the $j^{th}$ cluster of classes represented by their semantic vectors ($j \leq i$) when $i$ clusters are obtained. We run hierarchical agglomerative clustering (HAC) multiple times to decide the optimal number of clusters by evaluating the goodness of clusters in each $\Tilde{\Psi}_i$ as:
\begin{equation} \label{eq:7}
	N_z = \underset{2 \leq \, i \, \leq (N_s + N_{\Tilde{u}} - 1) }{argmax}\; 
	MSC(\Tilde{\Psi}_i)
\end{equation}
where $MSC(.)$ is the {\it mean silhouette coefficient}~\cite{ROUSSEEUW198753,kaufman2009finding}:
\begin{equation}
	\label{eq:8}
	MSC(\Tilde{\Psi}_i) =     \frac{1}{(N_s + N_{\Tilde{u}})}  \sum_{k \in \Tilde{\mathcal{C}}} \frac{b_k^i - a_k^i}{max \{a_k^i, b_k^i \}} 
\end{equation}
where $a_k^i$ and $b_k^i$ are the mean intra-cluster distance and mean nearest-cluster distance for semantic vector of class $k$ when $i$ clusters are formed by HAC. From the optimal set of clusters($\Tilde{\Psi}_{N_z}$), a single representative is selected from each $\Psi_{N_z}^j$ based on information from the cluster-specific semantic space.
\begin{figure}[t]
	\centering
	\includegraphics[width=\linewidth]{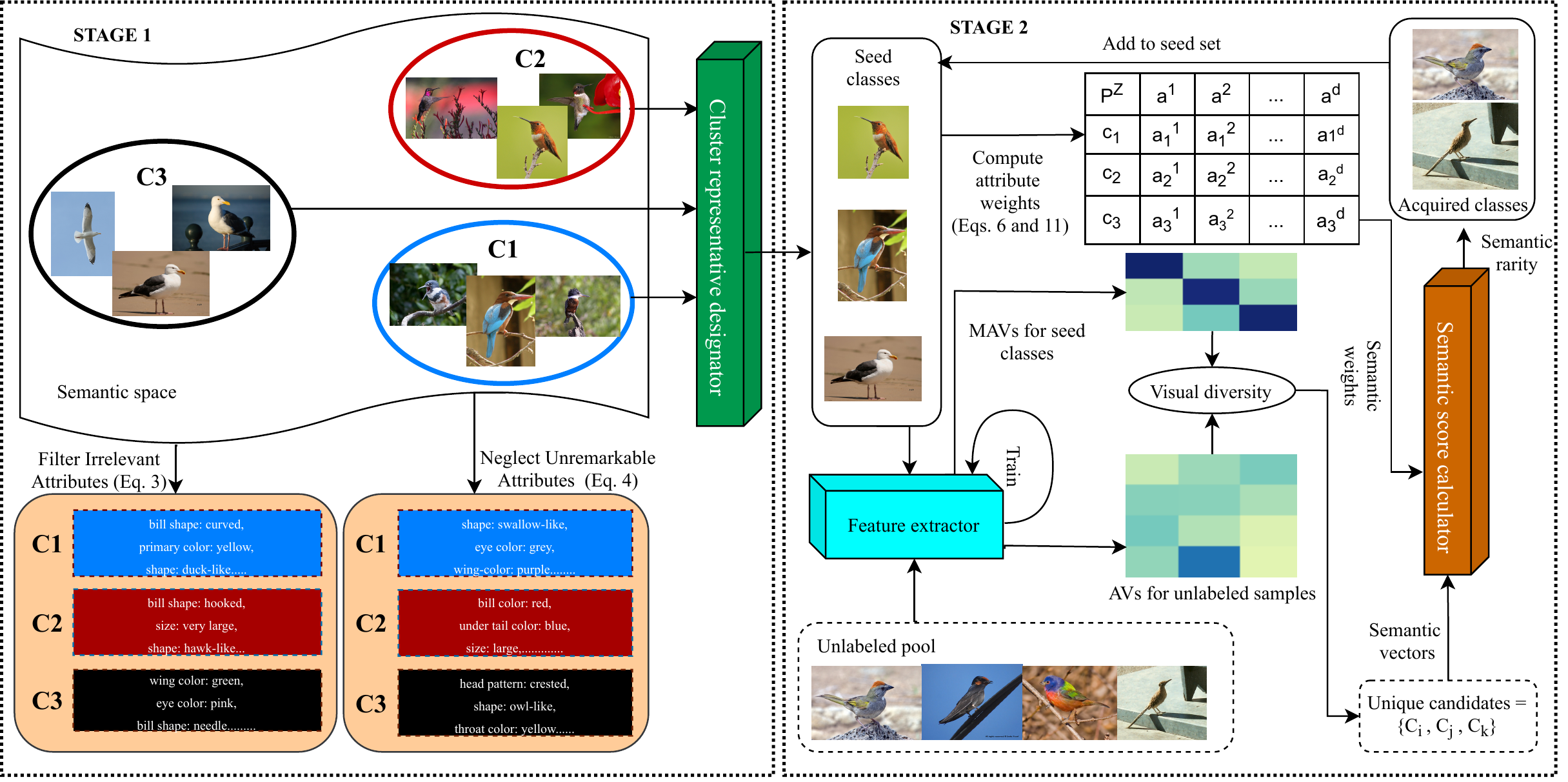}
	\caption{{\bf DiRaC-I workflow.} In stage 1, clustering in the semantic space selects a representative from each cluster while filtering out irrelevant and unremarkable attributes, and this forms the seed set. In stage 2, the seed set expands iteratively while considering visual diversity and semantic rarity, until it contains a fixed number of classes. The resulting set can act as seen set for training ZSL models}
	\Description{Two stages of our framework accounting for visual diversity and acknowledging semantic rarity in the object domain.}
	\label{fig:dirac-i}
	
\end{figure}
However, for a cluster, some attributes might not be present at all ({\bf irrelevant}) or may occur in minimal amounts ({\bf unremarkable}) and hence can be ignored while searching for its suitable representative. Therefore, for a cluster $\Psi_{N_z}^j$, we formally recognize these two groups of attributes respectively from the semantic space ($\mathcal{P}({\Psi_{N_z}^j})$) spanned by its member classes:
\begin{equation} \label{eq:9}
	IA(\Psi_{N_z}^j) = \{ a^l \in \mathbb{R} \mid a_c^l = 0, \; \forall \; c \in \Psi_{N_z}^j \}
\end{equation}
\begin{equation} \label{eq:10}
	UA(\Psi_{N_z}^j) = \{ a^l \in \mathbb{R} \mid \mathcal{B}_{c}^l(\Psi_{N_z}^j) = 0, \; \forall \; c \in \Psi_{N_z}^j \}
\end{equation}
where for an attribute $a^l$:
\begin{equation}
	\label{eq:11}
	\mathcal{B}_{c}^l(\Psi_{N_z}^j) = 
	\begin{cases}
		0,        &\text{if $a_c^l \leq \frac{1}{| \{c \in \Psi_{N_z}^j | \, a_c^l \neq 0 \} |} \sum \limits_{c \in \Psi_{N_z}^j} a_c^l  $} \\
		1,        &\text{otherwise}
	\end{cases}
\end{equation}
Here, for cluster $\Psi_{N_z}^j$, $IA(.)$ and $UA(.)$ denote the sets of irrelevant and unremarkable attributes respectively, and $\mathcal{B}(.)$ is a binary class-attribute matrix procured from $\mathcal{P}(.)$ after ignoring the irrelevant attributes. Unremarkable attributes are also discarded from both $\mathcal{B}(.)$ and $\mathcal{P}(.)$. To account for the rarity in semantic space $\mathcal{P}({\Psi_{N_z}^j})$, we calculate per-attribute frequencies with the help of the corresponding $\mathcal{B}(\Psi_{N_z}^j)$ --- rarer the attribute, more the importance given to it by sampling weights from the function:
\begin{equation} \label{eq:1}
	f(\theta_{a^l}) = - \log (\theta_{a^l})
\end{equation}
where we obtain attribute frequencies from the diagonal values ($d_l^l$) of matrix $(\mathcal{B}(\Psi_{N_z}^j))^T \cdot \mathcal{B}(\Psi_{N_z}^j)$ as:
\begin{equation}
	\label{eq:6}
	\theta_{a^l} = \frac{d_l^l}{| \Psi_{N_z}^j |}
\end{equation}
According to Eq.~\ref{eq:6}, $\theta \in (0, 1]$ and $f(\theta_{a^l}) \in [0, \infty)$. We use $- \log (\theta_{a^l})$ to sample attribute-weights as it is strictly decreasing on the interval $(0, 1]$, providing a higher weight if $a^l$ is rare (i.e., $\theta_{a^l}$ is low), and a lower weight otherwise. Finally, we get the seed set as $\mathcal{Z} = \{ \kappa(\Psi_{N_z}^1), \kappa(\Psi_{N_z}^2),...\kappa(\Psi_{N_z}^{N_z}) \}$ in which a representative class from each cluster is selected as:
\begin{equation} \label{eq:2}
	\kappa(\Psi_{N_z}^j) = \underset{c \in \Psi_{N_z}^j}{argmax}\;  (\mathcal{B}(\Psi_{N_z}^j)\odot \mathcal{P}(\Psi_{N_z}^j)) \cdot \mathcal{W}
\end{equation}
where $\odot$ denotes element-wise matrix multiplication and $\mathcal{W}$ is a vector of weights for attributes present in $\mathcal{P}(.)$. Such representatives from different clusters boost diversity while promoting semantic rarity via Eqs.~\ref{eq:1} and~\ref{eq:2}. We consider an outlier class (not a member of any cluster) to be diverse enough from the other classes and take it directly into the seed set.

\subsection{Stage 2: Visual-Semantic Mining (VSM)}
\label{sec:vsm}
The samples belonging to classes from $\mathcal{Z}$ act as a labeled data set used to initialize our VSM algorithm. VSM is inspired by Active Learning (AL), where inputs from an {\it Oracle} (the source of ground truth labels, e.g. a human expert) are used to label some of the most informative samples from an unlabeled pool ($\mathcal{A}$) to train machine learning models. For our framework, this pool corresponds to the samples not belonging to classes from the seed set (for a given VSM iteration only; dataset constructors actually provide labels for all samples available to DiRaC-I). We aim to adopt a similar strategy to iteratively acquire $N_s$ classes exhibiting diversity and rarity for training ZSL models.

In each iteration of VSM, we retrain a ResNet-101 ($\mathcal{M}$)~\cite{he2016deep} pretrained on ImageNet~\cite{russakovsky2015imagenet} to behave as a feature extractor for the seed class samples using a transfer learning approach. We capitalize on the work done by~\cite{bendale2016towards} and use the scores from the penultimate layer of a CNN (Activation Vectors or AVs) to estimate the distribution of the {\it related} classes, establishing a relationship between the unlabeled and labeled samples in the AV space. Each class $c \in \mathcal{Z}$ is represented by its {\bf Mean Activation Vector (MAV)} computed using the AVs of the training samples classified correctly by $\mathcal{M}$, obtaining the MAV set $\mathcal{V} = \{ \mu_1, \mu_2,...\mu_{| \mathcal{Z} |} \}$. For the unlabeled samples, we extract the AVs using the trained $\mathcal{M}$ to obtain $\mathcal{F} = \{ f_1, f_2,...f_{| \mathcal{A} |} \}$. We intend to capture the visually most diverse samples leveraging the AV space by first obtaining the set:
\begin{equation}
	\label{eq:12}
	\Pi = \{ k \in \mathbb{R} \mid k = \underset{\mu_c \in \mathcal{V}}{min} \; \delta(\mu_c, f_j), \; \forall f_j \in \mathcal{F} \}
\end{equation}
and then selecting $t$ samples from $\mathcal{A}$ corresponding to the largest values in $\Pi$, where $\delta$ denotes the Euclidean-cosine distance~\cite{bendale2016towards}. A set of unique {\it candidate classes} ($\mathcal{H}$) is formed by querying the ground truths of these $t$ samples. VSM then explores the rarity in the semantic space spanned by these candidate classes ($\mathcal{P}(\mathcal{H})$). A class-wise estimate of the number of images from the seed classes exhibiting each attribute can be obtained in a matrix $\mathcal{I}$, where:
\begin{equation}
	\label{eq:13}
	\mathcal{I}_c^l = a_c^l \cdot IC_c, \; c \in \mathcal{Z}
\end{equation}
Here, $a_c^l$ is an element from the semantic space of the seed classes ($\mathcal{P}(\mathcal{Z})$) and $IC_c$ gives the number of images for seed class $c$. New attribute-weights are calculated based on the proportion of each attribute within the currently ``known object domain ($\mathcal{Z}$)'' for VSM using Eq.~\ref{eq:1}, except that now:
\begin{equation}
	\label{eq:4}
	\theta_{a^l} = \frac{\sum \limits_{c \in \mathcal{Z}} \mathcal{I}_c^l}{\sum \limits_{c \in \mathcal{Z}} IC_c}
\end{equation}
Finally, we calculate {\bf semantic scores} for each candidate class as follows:
\begin{equation}
	\alpha_h = \mathcal{P}(\mathcal{H}) \cdot \mathcal{W}
	\label{eq:5}
\end{equation}
where $\mathcal{W}$ denotes the vector of obtained attribute-weights. Top-$q$ candidate classes having the highest semantic scores are added to $\mathcal{Z}$. Ground truth of the samples from the added classes are also queried at the end of an iteration so that samples belonging to classes in $\mathcal{Z}$ always remain labeled. We repeat this entire process (Fig.~\ref{fig:dirac-i}) until $| \mathcal{Z} | = N_s$ (kept as the same value as in Existing Split for a fair comparison).

It is important to note that VSM is {\bf only inspired by} Active Learning (AL). We do acknowledge the structural similarities with AL, such as seeds and acquisition functions. However, the problem setup and goal of VSM are quite different from AL. AL theoretically aims to select the most informative samples from a huge unlabeled pool of data, whereas that is not the case for DiRaC-I’s target group (dataset constructors). DiRaC-I can query the labels once a class is added to the seed set since labeled data for all the collected classes are available with the constructors. Hence, for this task, the AL assumption (model should not have access to labels) does not hold, and its absence does not make VSM impractical to use.

\section{Experiments} \label{expt}

\subsection{Datasets and seen-unseen splits}
\label{sec:5.1}
{\bf CUB}~\cite{WahCUB_200_2011} and {\bf SUN}~\cite{patterson2014sun} are two challenging fine-grained data sets, both with several classes but limited data per class. CUB contains 11,788 images from 200 bird categories, each of which is defined using 312 human-annotated attributes. SUN contains 14,340 images from 717 scene categories annotated with 102 attributes. $N_s$ is 150 for CUB and 645 for SUN for both ES and PS. Moreover, before initiating DiRaC-I, we obtain $N_{u_{com}}$ as 25 and 36 for CUB and SUN. The random split of $\mathcal{U}\textsubscript{E}$ is done three times and model evaluation is done on three different $\mathcal{U}_{com}$ sets --- $\mathcal{U}_{com}^1$, $\mathcal{U}_{com}^2$ and $\mathcal{U}_{com}^3$ --- where $X$ in $\mathcal{U}_{com}^X$ denotes the split number. Consequently, DiRaC-I runs three times with different object domains ($\Tilde{\mathcal{C}}$) at its inception. We report the image count of classes belonging to sets $\mathcal{S}\textsubscript{E}$, $\mathcal{S}\textsubscript{P}$ and $\mathcal{U}_{com}$ in Tab.~\ref{tab:1}, where the slight difference in image count for ES and PS can be attributed to the different seen classes considered in ES and PS. For visual features, we follow previous work~\cite{xian2018zero} and use CNN features extracted from pretrained ResNet-101~\cite{he2016deep}.

\subsection{Implementation details}
\label{sec:5.2}
During stage 1, HAC uses {\it Ward's method}~\cite{ward1963hierarchical} to calculate cluster similarity. Obtaining too few clusters (and hence, seed classes) using HAC would initialize the deep model ($\mathcal{M}$) in VSM with too few training samples. Additionally, some data sets have very few images per class --- e.g. 20 for SUN. Therefore, we set the lower bound of number of clusters to be formed as 5 to achieve effective model training. While retraining $\mathcal{M}$, weights of all the layers are frozen except the last fully-connected layer. The learning rates for optimizing $\mathcal{M}$ are set to 0.01 and 0.001 for CUB and SUN, respectively. $q$ is set to 2 for CUB and 4 for SUN. We need $t$ to be low so that in a practical scenario, only a small percentage of the unlabeled images from $\mathcal{A}$ need to be queried for their class labels while inferring the candidate classes during the entire process of VSM. In our experiments, $t = \max (5, \lceil(3 \log a) \rceil)$, where $a = $ average number of images per class for a given data set, according to which $t = 13$ for CUB and $t = 9$ for SUN. Across the three runs of DiRaC-I, VSM runs for 58 iterations on an average for CUB and queries the labels for 754 samples, i.e., 6.39\% of the total samples in CUB. For SUN, labels are queried for 10.04\% of the total samples (1440) over an average of 160 iterations. Furthermore, keeping in mind the real-life scenarios, we prioritize a candidate class to be included in $\mathcal{Z}$ if it is an overlapping class (Sec.~\ref{sec:3.2}), so that classes from the set $\Tilde{C} \setminus \mathcal{S}\textsubscript{P}$ can also serve as test set if required without violating zero-shot assumptions~\cite{xian2018zero}. We observe that these classes mostly have the highest semantic scores too, and hence conclude that the inclusion is fair.

\begin{table}[t]
	\begin{center}
		\caption{Image count in seen and common unseen classes for ES and PS for different random splits of $\mathcal{U}\textsubscript{E}$}
		\label{tab:1}      
		\begin{tabular}{lccccc}    
			\hline
			\multirow{2}{*}{\textbf{Dataset}}   & 
			\multirow{2}{*}{\textbf{Split}}  & \multicolumn{2}{c}{\textbf{ES}} & \multicolumn{2}{c}{\textbf{PS}}
			\\ 
			
			& & $\boldsymbol{\mathcal{S}\textsubscript{E}}$ & $\boldsymbol{\mathcal{U}_{com}}$ & $\boldsymbol{\mathcal{S}\textsubscript{P}}$ & $\boldsymbol{\mathcal{U}_{com}}$\\
			\hline
			
			\multirow{3}{*}{CUB} 
			&1 &7057 &1489 &7068 &1489 \\
			&2 &7057 &1488 &7068 &1488 \\
			&3 &7057 &1471 &7075 &1471 \\
			
			\hline
			
			\multirow{3}{*}{SUN} 
			&1 &10320 &720 &10320 &720 \\
			&2 &10320 &720 &10320 &720 \\
			&3 &10320 &720 &10320 &720 \\
			
			\hline
			
		\end{tabular}
	\end{center}
\end{table}

\begin{table}[t]
	\begin{center}
		\caption{Comparative results (top-1 accuracy in \%) of Conventional ZSL with models from sets $\Phi_E$ and $\Phi_P$ (defined in Sec.~\ref{sec:3.2}) on the CUB and SUN data sets. Results on test classes having at least one common attribute corresponds to results on all classes (left), since all classes in CUB and SUN exhibit at least one common attribute (see Tab.~\ref{tab:3}). Enhanced results achieved with PS are in {\bf BOLD}}
		\label{tab:2}    
		\begin{tabular}{lccccc|cccc}
			\hline
			\multirow{4}{*}{{\bf Method}}   & 
			\multirow{4}{*}{{\bf Test set}}  &
			\multicolumn{4}{c}{{\bf ZSL for all}} &
			\multicolumn{4}{c}{{\bf ZSL for classes having}} \\
			
			& & \multicolumn{4}{c}{{\bf test classes}} &
			\multicolumn{4}{c}{{\bf at least 1 rare attribute}} 
			\\ 
			
			& & \multicolumn{2}{c}{{\bf CUB}} & \multicolumn{2}{c}{{\bf SUN}} & \multicolumn{2}{c}{{\bf CUB}} &
			\multicolumn{2}{c}{{\bf SUN}} 
			\\
			& & {\bf ES} & {\bf PS} & {\bf ES} & {\bf PS} & {\bf ES\textsubscript{R}} & {\bf PS\textsubscript{R}} & {\bf ES\textsubscript{R}} & {\bf PS\textsubscript{R}} \\
			\hline
			
			\multirow{3}{*}{\textbf{ALE} \cite{akata2015label}} 
			&$\mathcal{U}_{com}^1$ &47.85 &\textbf{50.76} &61.25 &59.17 &46.58 &\textbf{49.89} &63.63 &60.45 \\
			&$\mathcal{U}_{com}^2$ &43.76 &\textbf{47.62} &63.19 &59.86 &41.32 &\textbf{48.36} &61.84 &58.68 \\
			&$\mathcal{U}_{com}^3$ &48.06 &\textbf{56.10} &60.28 &57.92 &47.99 &\textbf{54.24} &61.66 &\textbf{63.33} \\
			
			\hline

			\multirow{3}{*}{\textbf{SAE} \cite{kodirov2017semantic}} 
			&$\mathcal{U}_{com}^1$ &40.69 &\textbf{44.07} &47.78 &\textbf{53.19}  &39.26 &\textbf{42.85} &52.5 &\textbf{59.54}\\
			&$\mathcal{U}_{com}^2$ &32.07 &\textbf{39.64} &54.17 &\textbf{55.69} &30.30 &\textbf{40.12} &53.68 &\textbf{55.26}\\
			&$\mathcal{U}_{com}^3$ &40.41 &\textbf{43.64} &50.69 &\textbf{52.92} &40.42 &\textbf{44.11} &50.00 &\textbf{55.66}\\
			
			\hline
			
			\multirow{3}{*}{\textbf{SJE} \cite{akata2015evaluation}} 
			&$\mathcal{U}_{com}^1$ &48.01 &\textbf{50.74} &53.19 &53.06  &47.16 &\textbf{49.37} &54.31 &\textbf{55.90} \\
			&$\mathcal{U}_{com}^2$ &40.26 &\textbf{47.98} &53.19 &\textbf{55.42} &38.32 &\textbf{47.32} &51.57 &\textbf{56.57}\\
			&$\mathcal{U}_{com}^3$ &54.05 &\textbf{54.81} &51.67 &50.28 &48.20 &\textbf{48.72} &55.00 &53.66 \\
			
			\hline
			
			\multirow{3}{*}{\textbf{DeViSE} \cite{frome2013devise}} 
			&$\mathcal{U}_{com}^1$ &47.18 &\textbf{50.27} &53.89 &\textbf{54.31} &45.46 &\textbf{48.96} &57.04 &\textbf{60.22}\\
			&$\mathcal{U}_{com}^2$ &42.55 &\textbf{45.63} &58.47 &55.00  &38.88 &\textbf{45.56} &56.31 &56.31 \\
			&$\mathcal{U}_{com}^3$ &44.58 &\textbf{45.09 }&55.14 &54.17 &41.32 &\textbf{43.15} &62.00 &\textbf{66.00}\\
			
			\hline
			
			\multirow{3}{*}{\textbf{ESZSL} \cite{romera2015embarrassingly}} 
			&$\mathcal{U}_{com}^1$ &53.62 &52.37 &53.89 &49.31 &52.79 &51.01 &59.54 &56.81 \\
			&$\mathcal{U}_{com}^2$ &44.46 &\textbf{50.80} &55.42 &52.78 &44.38 &\textbf{52.27} &57.36 &53.15\\
			&$\mathcal{U}_{com}^3$ &56.81 &\textbf{59.84} &57.78 &50.83 &51.63 &\textbf{55.96} &62.66 &54.00\\
			
			\hline
			
			\multirow{3}{*}{\textbf{LsrGAN} \cite{vyas2020leveraging}} 
			&$\mathcal{U}_{com}^1$ &57.91 &\textbf{60.19} &59.44 &\textbf{64.58} &57.61 &\textbf{59.64} &62.72 &\textbf{65.90}\\
			&$\mathcal{U}_{com}^2$ &56.03 &\textbf{59.40} &61.67 &\textbf{64.17} &56.01 &\textbf{61.28} &61.57 &\textbf{63.94}\\
			&$\mathcal{U}_{com}^3$ &64.43 &61.46 &60.69 &\textbf{61.81} &60.54 &57.84 &63.66 &\textbf{66.00}\\
			
			\hline
			
			\multirow{3}{*}{{\bf TF-VAEGAN} \cite{narayan2020latent}} 
			&$\mathcal{U}_{com}^1$ &63.48 &\textbf{66.32} &64.58 &\textbf{65.83} &63.68 &\textbf{65.89} &65.00 &\textbf{67.04}\\
			&$\mathcal{U}_{com}^2$ &61.82 &\textbf{64.98} &69.03 &66.11 &60.55 &\textbf{65.13} &68.15 &67.10 \\
			&$\mathcal{U}_{com}^3$ &67.56 &\textbf{68.35} &65.83 &65.00 &64.44 &\textbf{65.06} &67.66 &\textbf{70.33} \\

			\hline
		\end{tabular}
	\end{center}
\end{table}

\subsection{Performance comparison with Existing Splits} \label{sec:5.3} 
We report the average per-class top-1 accuracy~\cite{xian2018zero} in the CZSL setting to evaluate ZSL methods based on different approaches like compatibility learning and generative frameworks and compare the performance of $\Phi_E$ and $\Phi_P$ (Sec.~\ref{sec:3.2}). All the methods are implemented in PyTorch. Among them, official codes in PyTorch are available for LsrGAN~\cite{vyas2020leveraging} and TF-VAEGAN~\cite{narayan2020latent}, and the rest are re-implemented versions based on the original publications. The hyperparameters for the official codes are directly used, whereas they are set on the validation sets for the rest. Note that model accuracy on ES reported in Tab.~\ref{tab:2} is not comparable with that of the original papers because the test set is different in our framework. Table~\ref{tab:2} shows that in the CZSL setting, models in $\Phi_P$ show significant improvements on CUB over $\Phi_E$ across all three splits. For SUN, we notice a mix of improved and similar results with models in $\Phi_E$. This might be because DiRaC-I leverages information from the semantic space, which lacks attributes with discriminative strength in the case of SUN, as explained in Sec.~\ref{sec:6.1}. Comparing results for ES and PS in the GZSL setting would not be fair in the current work because the seen classes in ES and PS might be different. Consequently, they might have different influences on the unseen predictions and the harmonic mean of seen and unseen accuracy (GZSL evaluation metric). This is true for most models because of their bias towards the seen classes.

\section{Framework analysis}\label{fr-ana}
In this section, we demonstrate some qualitative and quantitative results and analyze the performance of the two stages of our framework, as well as the impact of incorporating diversity and rarity of the object domain in the training of ZSL models. In some of the subsections that follow, we show some qualitative results on the {\bf CUB}~\cite{WahCUB_200_2011} data set when $\Tilde{\mathcal{C}} = \mathcal{S}\textsubscript{E} \, \cup \, (\mathcal{U}\textsubscript{E} \setminus \mathcal{U}_{com}^2)$ ($\mathcal{U}_{com}^2$ denotes the set $\mathcal{U}_{com}$ after the randomly splitting $\mathcal{U}\textsubscript{E}$ for the second time). For brevity, we denote this set as $\Tilde{\mathcal{C}^2}$.
All the results are shown for the object domain $\Tilde{\mathcal{C}^2}$ (unless stated otherwise) to maintain a correlation between results obtained from various stages of the framework. We choose the CUB dataset for qualitative results because the attributes that characterize each class in CUB are visually interpretable and hence can be easily verified with the visual results we provide here.

\begin{figure}[t]
	\centering
	\includegraphics[width=\linewidth]{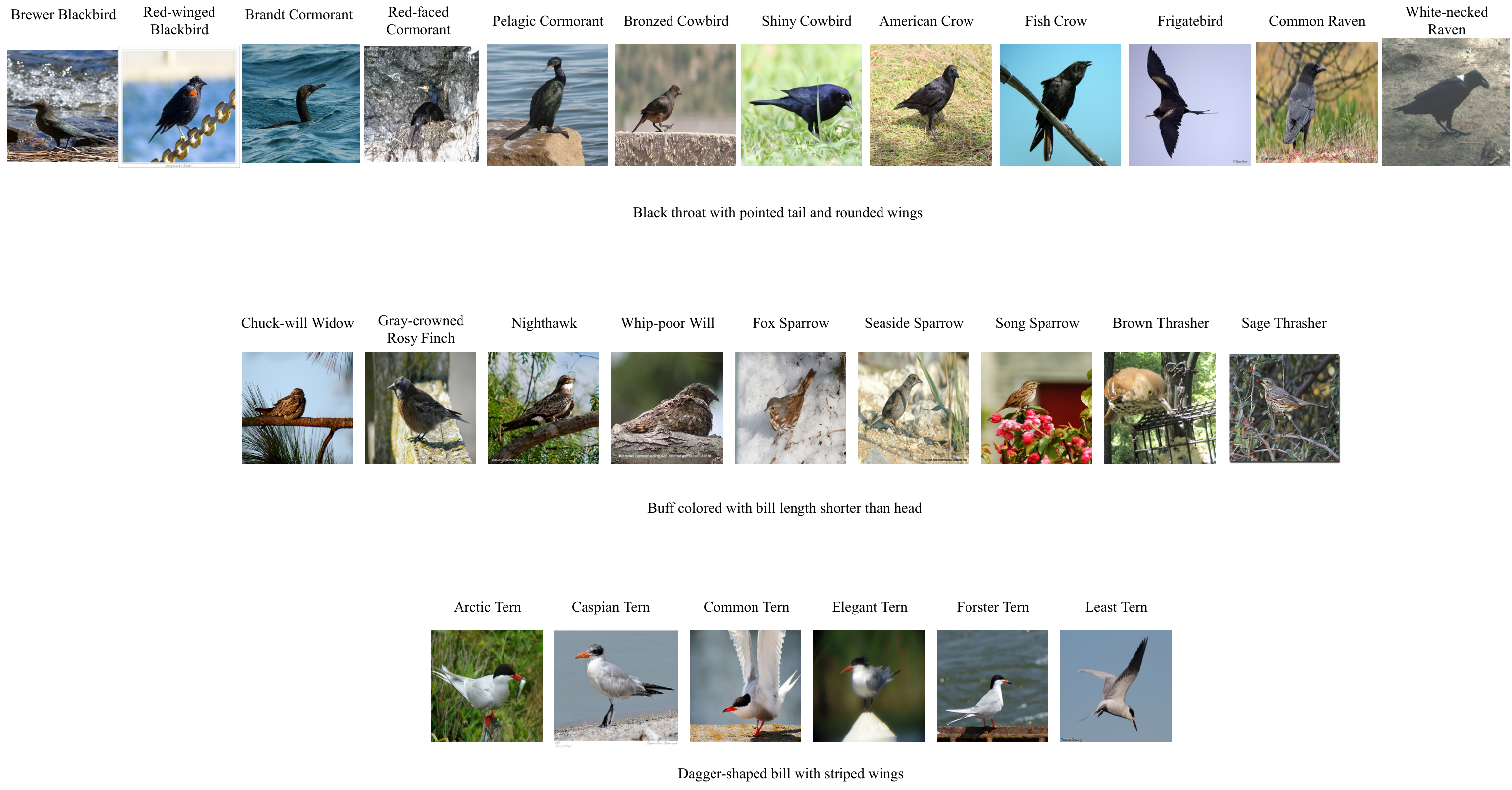}
	
	\caption{Results of clustering during seed-set construction (for object domain $\Tilde{\mathcal{C}^2}$). Each row corresponds to the members of a specific cluster obtained. The attribute descriptions for each cluster written below each row are formed by combining some of the most frequent attributes, procured after discarding the attributes adjudged as \textit{irrelevant} and \textit{unremarkable} for the cluster (such discarded attributes can be found in Tab.~\ref{tab:ia_ua}) }
	\Description{Verifying the visual images within clusters with their corresponding semantic descriptions}
	\label{fig:clusters_with_att}
\end{figure}

\subsection{Seed-set construction} \label{sec:6.1} 
We aim to achieve a good quality of clusters during this stage and acquire seed classes that provide a comprehensive initial idea of the object domain to the next stage of DiRaC-I. For CUB, obtained clusters are visually more interpretable since the semantic space consists of several {\it groups} of discriminative properties like {\it wing color, bill shape, head pattern}, etc. Hence, {\it hummingbirds, kingfishers, and gulls} get clustered separately, and picking a representative from each cluster captures the object domain diversity well enough. However, the SUN attributes come from a variety of contexts~\cite{patterson2014sun}, many of which are applicable to several classes with the same attribute strength --- e.g. $\mathcal{P}(\mathcal{S}\textsubscript{E} \cup (\mathcal{U}\textsubscript{E} \setminus \mathcal{U}_{com}^1))$ shows attributes {\it warm} and {\it eating} have a non-zero value for 669 and 346 classes but have only 50 and 49 unique values. This results in a large number of classes clustering together in the semantic space (Fig.~\ref{fig:semantic_vsm}). Hence, our experiments suggest that data sets which characterize classes using more discriminative attribute strengths would help in selecting better seed classes. Figures~\ref{fig:semantic1_CUB} and~\ref{fig:semantic1_SUN} show the seed classes (numbered in `black') for CUB and SUN respectively. Randomly selecting such classes could pick all of them from a particular region (when every class is semantically very similar) or from very different regions. However, our approach leverages the semantic relationships between classes and ensures that it picks diverse representatives, as evident from Fig.~\ref{fig:semantic_vsm}.

\begin{figure}[t]
	\centering
	\includegraphics[width=0.5\linewidth]{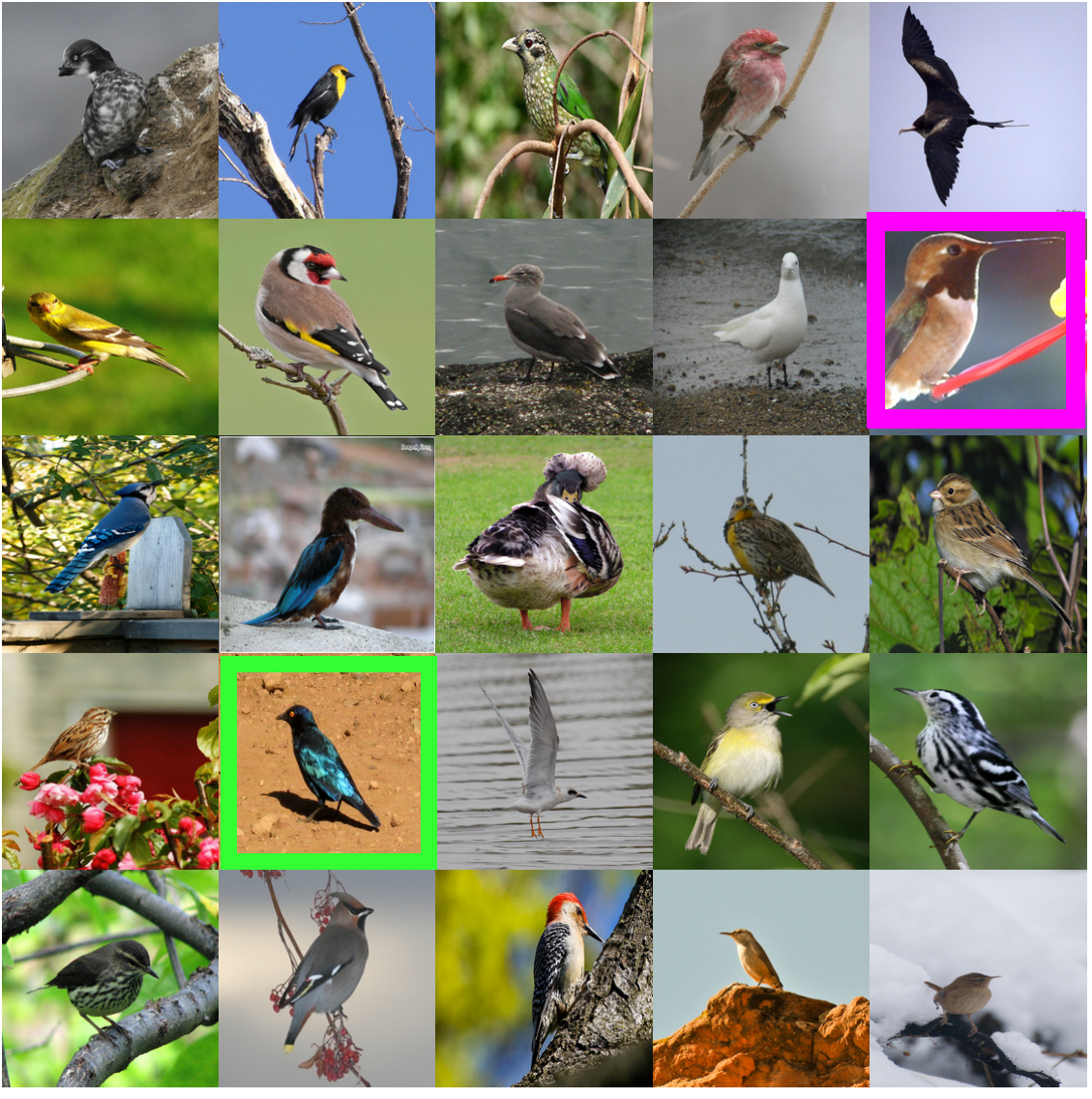}
	
	\caption{An example from every seed class at the end of stage 1 for CUB with object domain $\Tilde{\mathcal{C}^2}$. Notice the birds exhibiting rare attributes of the domain $\Tilde{\mathcal{C}^2}$ (see Tab.~\ref{tab:rare_common}) such as needle-shaped bill (pink box) and orange eye (green box)}
	\Description{Prioritizing rare attributes to identify object domain}
	\label{fig:seeds0}
\end{figure}

\begin{table}[t]
	\begin{center}
		\caption{Sets of irrelevant attributes ($IA$) and unremarkable attributes ($UA$) obtained for the clusters corresponding to the top, middle and bottom rows of Fig.~\ref{fig:clusters_with_att}, respectively. $\Psi^c_{25}$ denotes cluster number $c$ out of the 25 clusters obtained using HAC. (+$p$) indicates set contains $p$ more attributes}
		\label{tab:ia_ua}
		\begin{tabular}{cll}
			\hline
			\multirow{1}{*}{{\bf Cluster (c)}}   & 
			\multirow{1}{*}{\bf $\boldsymbol{IA(\Psi_{25}^c)}$}   & 
			\multirow{1}{*}{\bf $\boldsymbol{UA(\Psi_{25}^c)}$}  \\

			\hline
			
			\multirow{5}{*}{1} 
			&Purple bill &Yellow back \\
			&Olive bill &Purple eye \\
			&Green bill &Pink forehead \\
			&Rufous crown &Orange nape \\
			& &Green leg (+14) \\
			\hline
			
			\multirow{5}{*}{2} 
			&Spatulate-shaped bill &Purple wing \\
			&Green wing &Green throat \\
			&Purple leg &Blue under-tail \\
			&Olive crown &Red nape \\
			&Pink eye (+19) &Blue belly (+24) \\
			\hline
			
			\multirow{5}{*}{3} 
			&Purple back &Orange wing \\
			&Pink underparts &Red upper-tail \\
			&Green upper-tail &Brown eye \\
			&Red forehead &Blue nape \\
			&Olive breast (+57) &Pink bill (+53) \\

			\hline
		\end{tabular}
	\end{center}
\end{table}

\begin{figure}[t]
	\centering
	\includegraphics[width=0.45\linewidth]{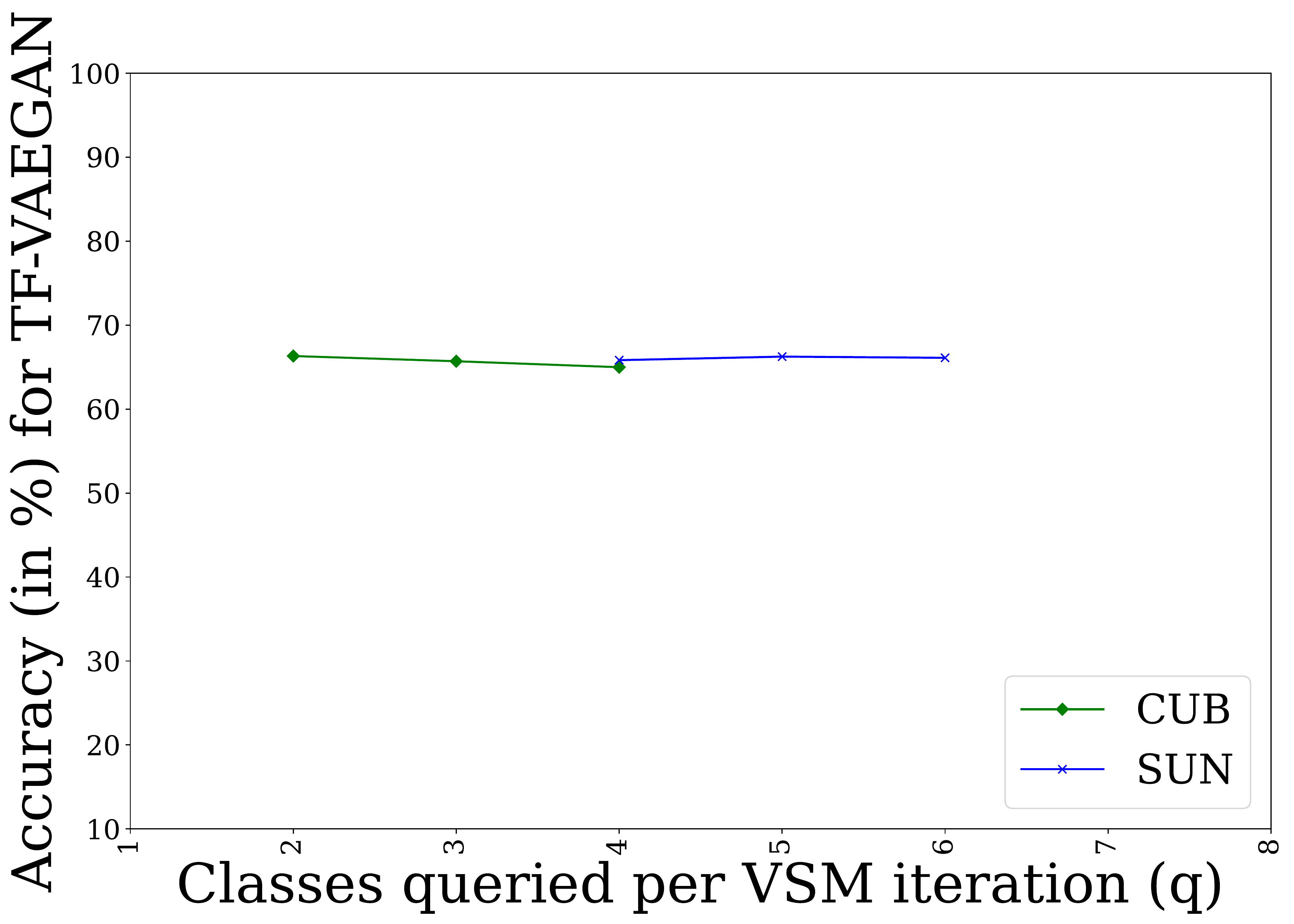}
	\caption{Sensitivity to $q$ for TF-VAEGAN~\cite{narayan2020latent}}
	\Description{Number of classes selected each time do not hamper model accuracy too much}
	\label{fig:hyperparam}
\end{figure}

\subsubsection{Clustering in the semantic space} \label{sec:seed.1}
\begin{figure}[t]
	\centering
	\begin{subfigure}[t]{0.4\textwidth}
		\includegraphics[scale=0.2]{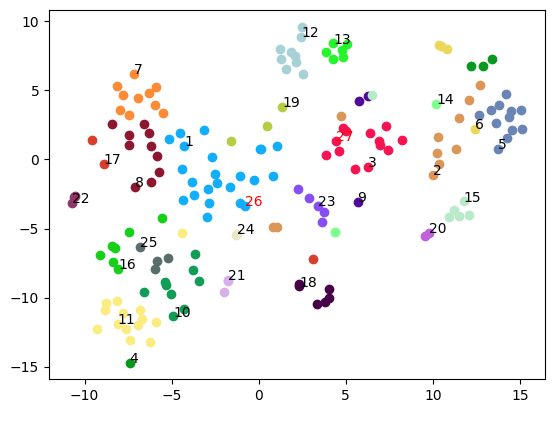}
		\caption{CUB, $i_1$}
		\label{fig:semantic1_CUB}
	\end{subfigure} \hfil
	\begin{subfigure}[t]{0.4\textwidth}
		\includegraphics[scale=0.2]{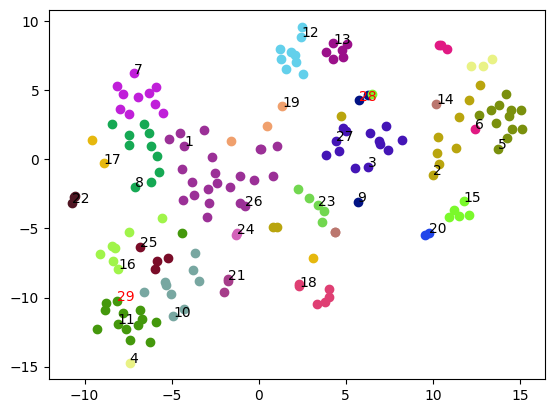}
		\caption{CUB, $i_2$}
		\label{fig:semantic2_CUB}
	\end{subfigure} \hfil
	\begin{subfigure}[t]{0.4\textwidth}
		\includegraphics[scale=0.2]{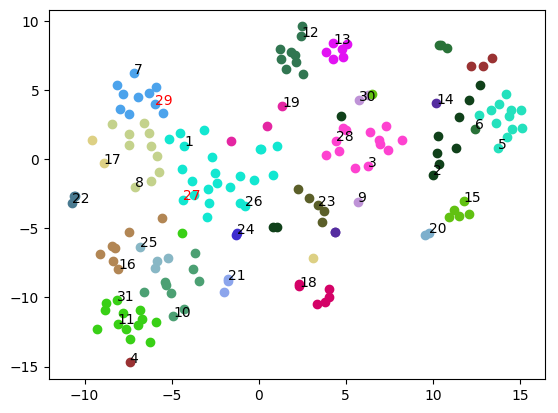}
		\caption{CUB, $i_3$}
		\label{fig:semantic3_CUB}
	\end{subfigure} \hfil
	\begin{subfigure}[t]{0.4\textwidth}
		\includegraphics[scale=0.2]{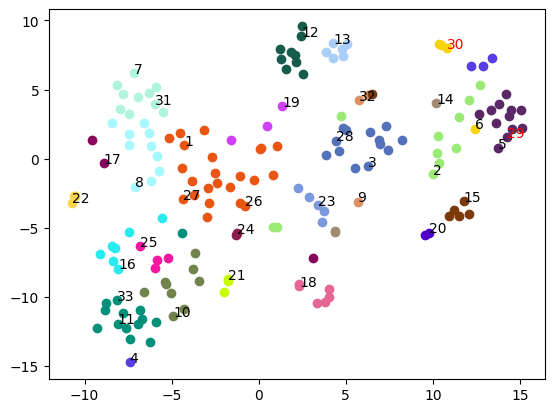}
		\caption{CUB, $i_4$}
		\label{fig:semantic4_CUB}
	\end{subfigure}
	
	\begin{subfigure}[t]{0.4\linewidth}
		\includegraphics[scale=0.2]{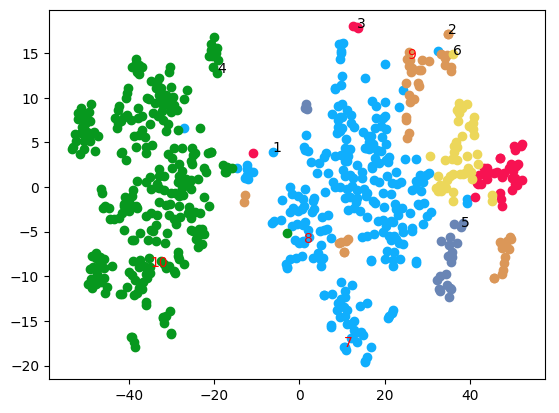}
		\caption{SUN, $i_1$}
		\label{fig:semantic1_SUN}
	\end{subfigure}
	\hfil
	\begin{subfigure}[t]{0.4\linewidth}
		\includegraphics[scale=0.2]{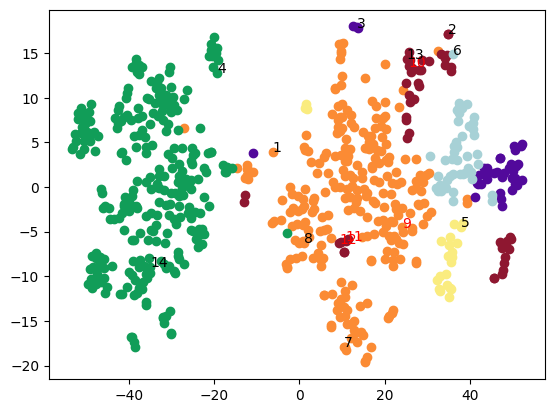}
		\caption{SUN, $i_2$}
		\label{fig:semantic2_SUN}
	\end{subfigure}
	\hfil
	\begin{subfigure}[t]{0.4\linewidth}
		\includegraphics[scale=0.2]{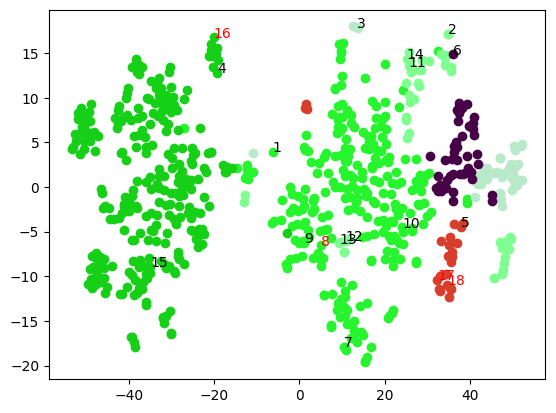}
		\caption{SUN, $i_3$}
		\label{fig:semantic3_SUN}
	\end{subfigure}
	\hfil
	\begin{subfigure}[t]{0.4\linewidth}
		\includegraphics[scale=0.2]{latex/u_split1_SUN_tsne_clusters_q2.png}
		\caption{SUN, $i_4$}
		\label{fig:semantic4_SUN}
	\end{subfigure}
	\caption{Visualization of the classes in the semantic space acquired during different iterations of VSM for both CUB and SUN by t-SNE method~\cite{van2014accelerating} (best viewed in color). Each class is represented by its attribute vector in the semantic space, and classes in the same cluster are shown in the same color. The top $q$ classes acquired in the $k^{th}$ iteration are marked with `red' numbers, and the rest of the numbered classes are the seed classes before starting iteration named $i_k$}
	\Description{A representation of how classes spanning various regions of the semantic space are captured in each VSM iteration}
	\label{fig:semantic_vsm}
\end{figure}

Hierarchical Agglomerative Clustering (HAC) in the semantic space spanned by classes from  set $\Tilde{\mathcal{C}^2}$ provides 25 clusters, and a single representative is designated as the seed class from each cluster. Fig.~\ref{fig:clusters_with_att} elucidates the clustering quality by presenting the cluster members of three clusters found by HAC. It can be seen that all the cluster members of the bottom row belong to the family of {\it terns} --- hence picking a seed class from this cluster ensures that the final seed set at the end of seed-set construction has a member from this family of birds. For the other rows, although the cluster members come from several families, they share common visual properties. For example, {\it crows, cormorants, blackbirds} and others have been clustered together in the top row, whereas the middle row consists of small-sized birds like {\it sparrows, finches,} etc. This suggests that selecting a member from each cluster would provide a good visual representation of that cluster to the next stage (VSM). Fig.~\ref{fig:seeds0} shows a sample image from each of the seed classes of CUB dataset corresponding to Fig.~\ref{fig:semantic1_CUB} with object domain $\Tilde{\mathcal{C}^2}$, providing a visual idea of the domain diversity captured within the seed classes.

\subsubsection{Designating cluster representatives as seeds}
Once the clusters are obtained, seed classes are selected based on the cluster-specific semantic space using Eqs.~\ref{eq:1},~\ref{eq:6}, and~\ref{eq:2}. However, to ensure that the computations are devoid of the effects of {\it irrelevant} and {\it unremarkable} attributes of a cluster, we defined sets $IA(.)$ and $UA(.)$ for every cluster. Table~\ref{tab:ia_ua} shows these two sets obtained corresponding to the three clusters (out of 25) exhibited in Fig.~\ref{fig:clusters_with_att}. Combining the information from Fig.~\ref{fig:clusters_with_att} and Tab.~\ref{tab:ia_ua}, we can see that the attributes belonging to the obtained sets $IA(.)$ and $UA(.)$ are indeed not descriptive enough of the cluster members. However, associating some of the most frequently occurring attributes in a given cluster, we construct some cluster descriptions (Fig.~\ref{fig:clusters_with_att}) and find them to be consistent with the visual images of the cluster members, providing a general description of the cluster.


\setlength{\tabcolsep}{1.4pt}

\begin{figure}[t]
	\centering
	\begin{subfigure}{0.7\textwidth}
		\includegraphics[scale = 0.18]{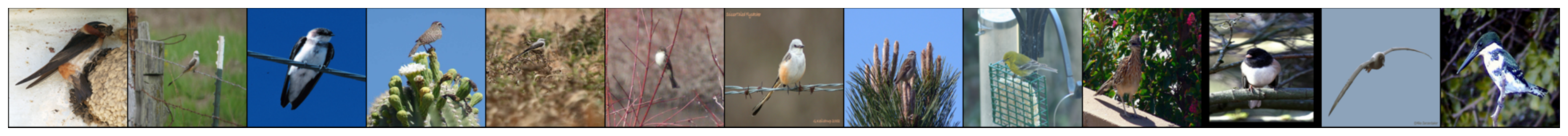}
		\caption{After iteration 1}
		\label{fig:vsm1}
	\end{subfigure}
	
	\begin{subfigure}{0.7\textwidth}
		\includegraphics[scale = 0.18]{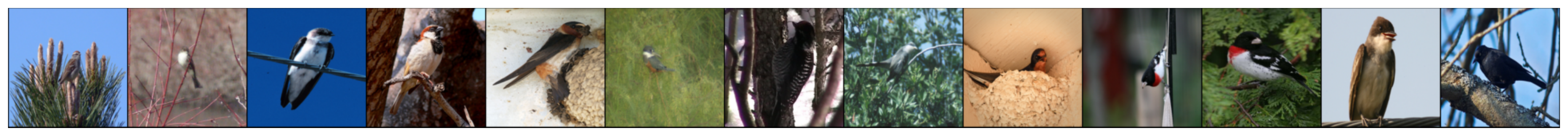}
		\caption{After iteration 2}
		\label{fig:vsm2}
	\end{subfigure}
	
	\begin{subfigure}{0.7\textwidth}
		\includegraphics[scale = 0.18]{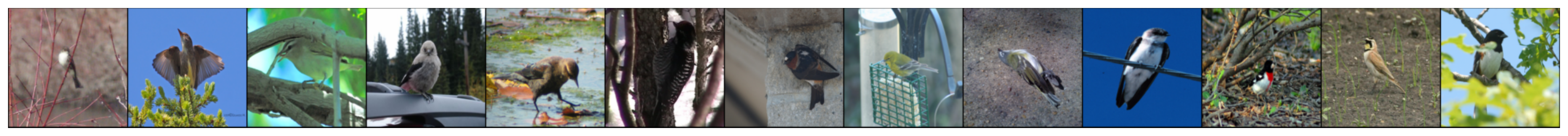}
		\caption{After iteration 3}
		\label{fig:vsm3}
	\end{subfigure}
	
	\begin{subfigure}{0.7\textwidth}
		\includegraphics[scale = 0.18]{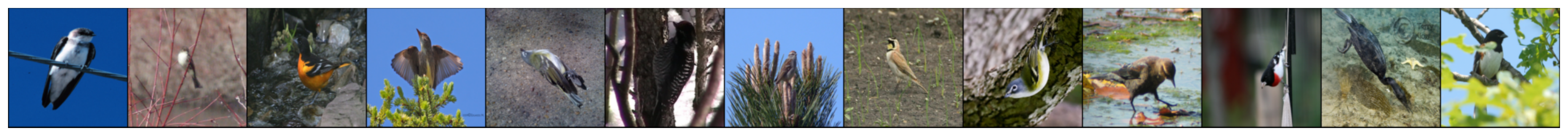}
		\caption{After iteration 4}
		\label{fig:vsm4}
	\end{subfigure}

	\caption{Top $t$ samples visually most diverse from the seed classes at the start of iteration $i$, based on a Euclidean-cosine distance~\cite{bendale2016towards}}
	\Description{Visual diversity captured within the object domain of birds}
	\label{fig:vsm_candidates}
\end{figure}

\subsection{Visual-Semantic Mining (VSM)} \label{sec:6.2} 
The idea of capturing diversity and rarity in the object domain via an iterative VSM algorithm is pivotal to our work and has been shown in action in Fig.~\ref{fig:semantic_vsm}. We notice that classes are captured from several regions of the semantic space, maximizing the distance from the existing seed classes in most cases. In a few cases, the acquired classes are closer to quite a few existing seed classes, like classes labeled as 29 in Fig.~\ref{fig:semantic4_CUB} and 8 in Fig.~\ref{fig:semantic3_SUN}. These cases arise when the generated semantic scores exceed the visual diversity factor (Eq.~\ref{eq:12}) by virtue of the rarity of attributes.  

\subsubsection{Qualitative analysis: VSM}
\label{sec:vsm-qual}
\begin{figure}
	\centering
	\begin{subfigure}{0.4\textwidth}
		\includegraphics[scale=0.4]{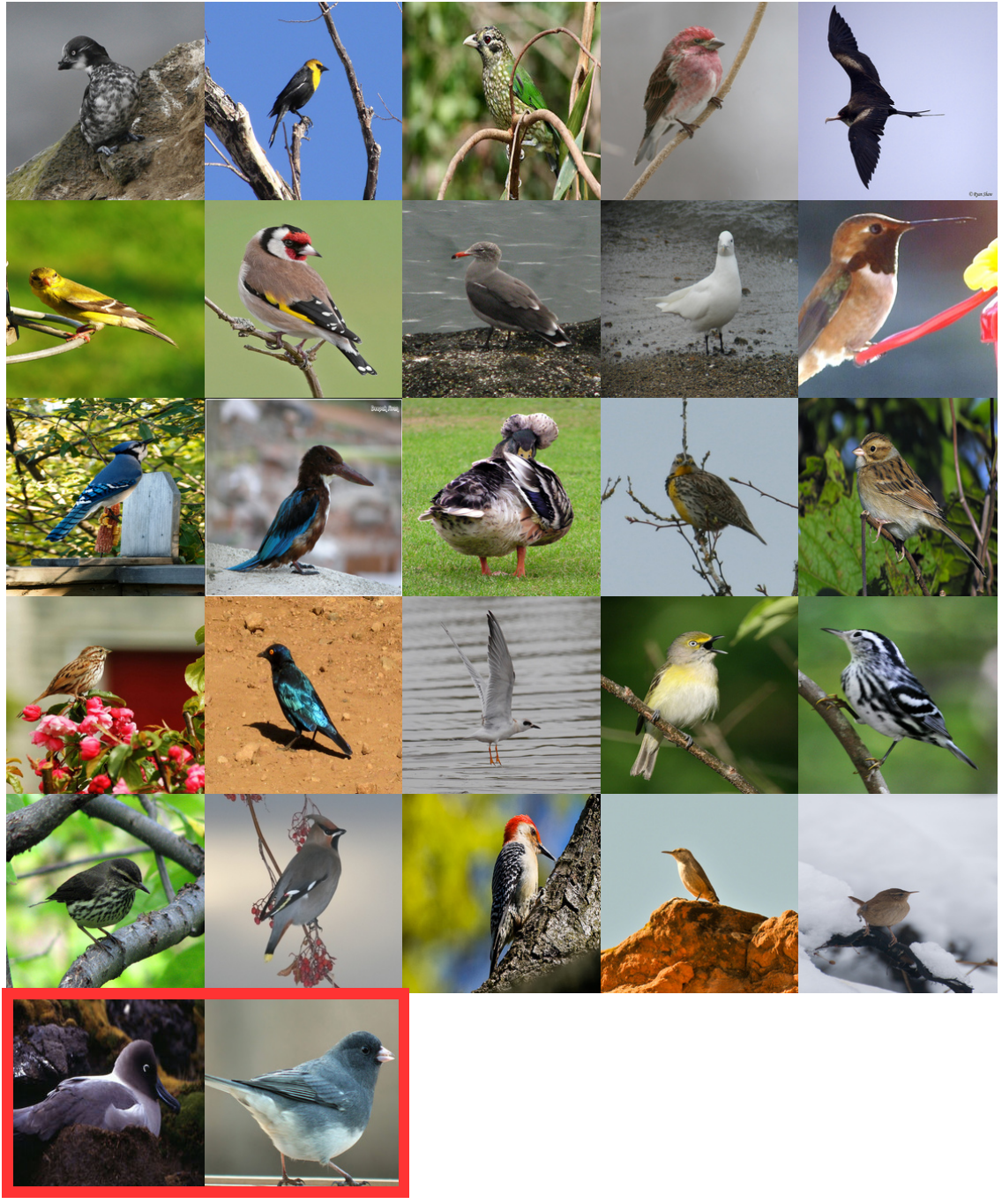}
		\caption{After iteration 1}
		\label{fig:seeds1}
	\end{subfigure}
	\hfil
	\begin{subfigure}{0.4\textwidth}
		\includegraphics[scale=0.4]{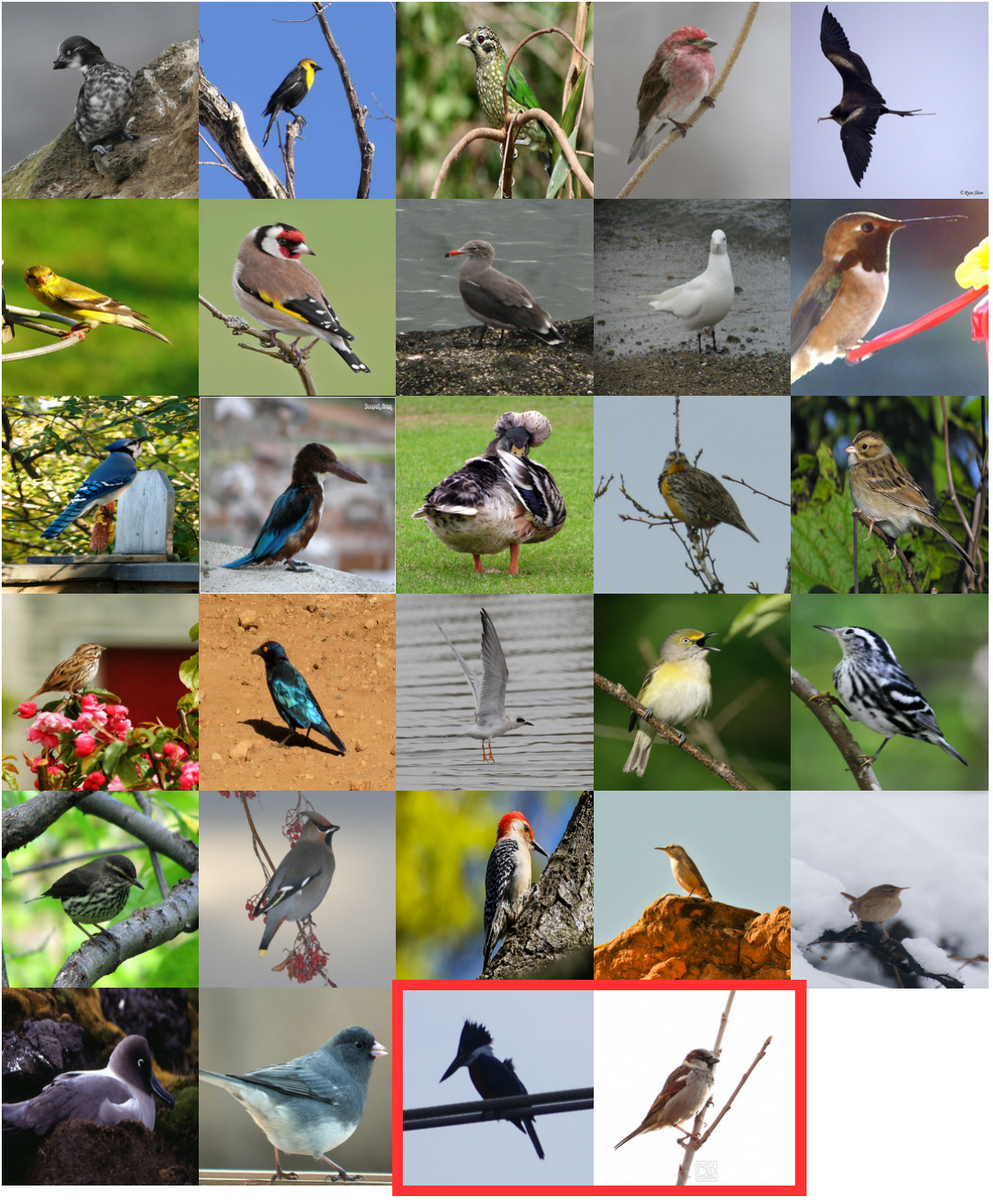}
		\caption{After iteration 2}
		\label{fig:seeds2}
	\end{subfigure}

	\begin{subfigure}{0.4\textwidth}
		\includegraphics[scale=0.4]{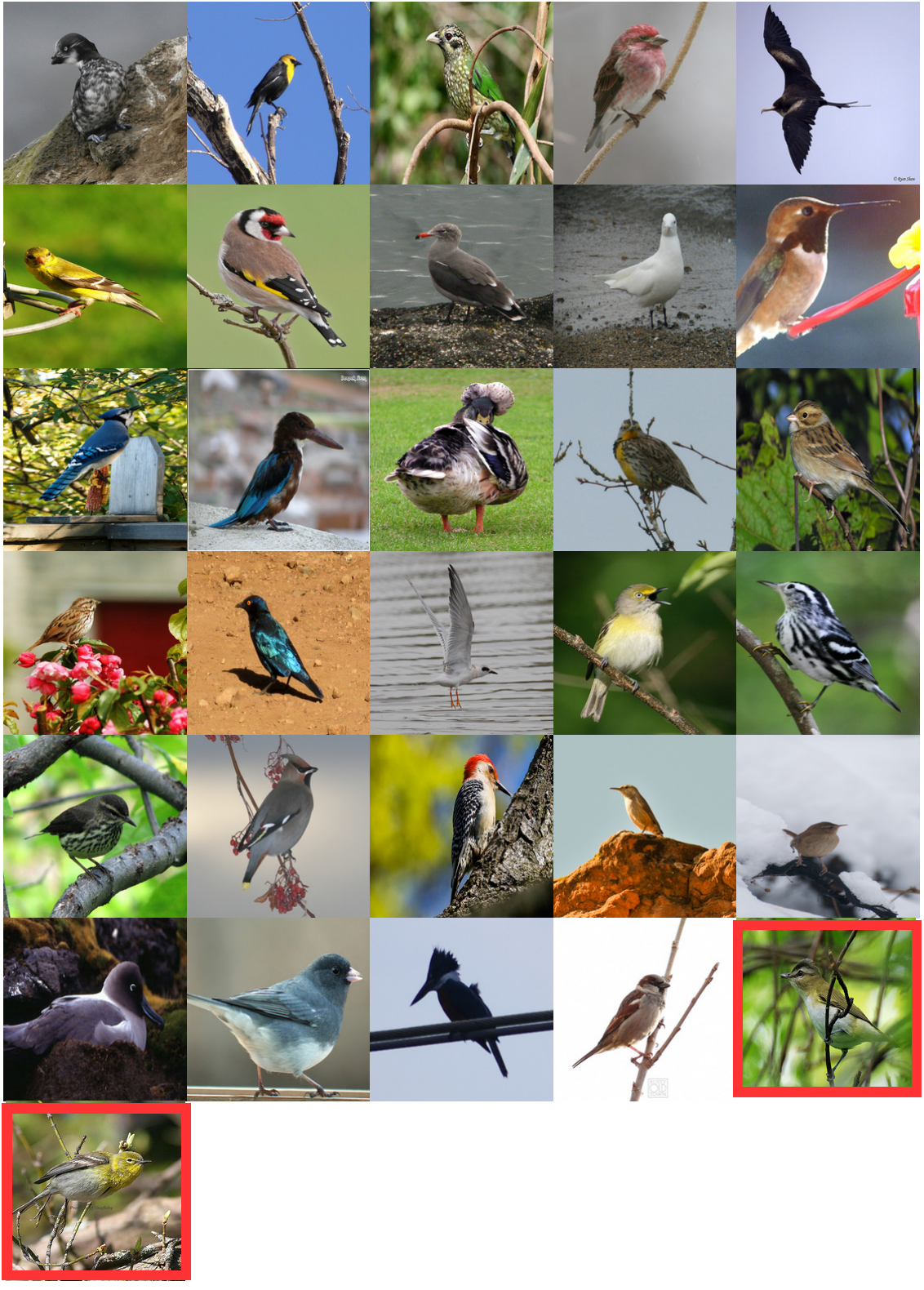}
		\caption{After iteration 3}
		\label{fig:seeds3}
	\end{subfigure}
	\hfil
	\begin{subfigure}{0.4\textwidth}
		\includegraphics[scale=0.4]{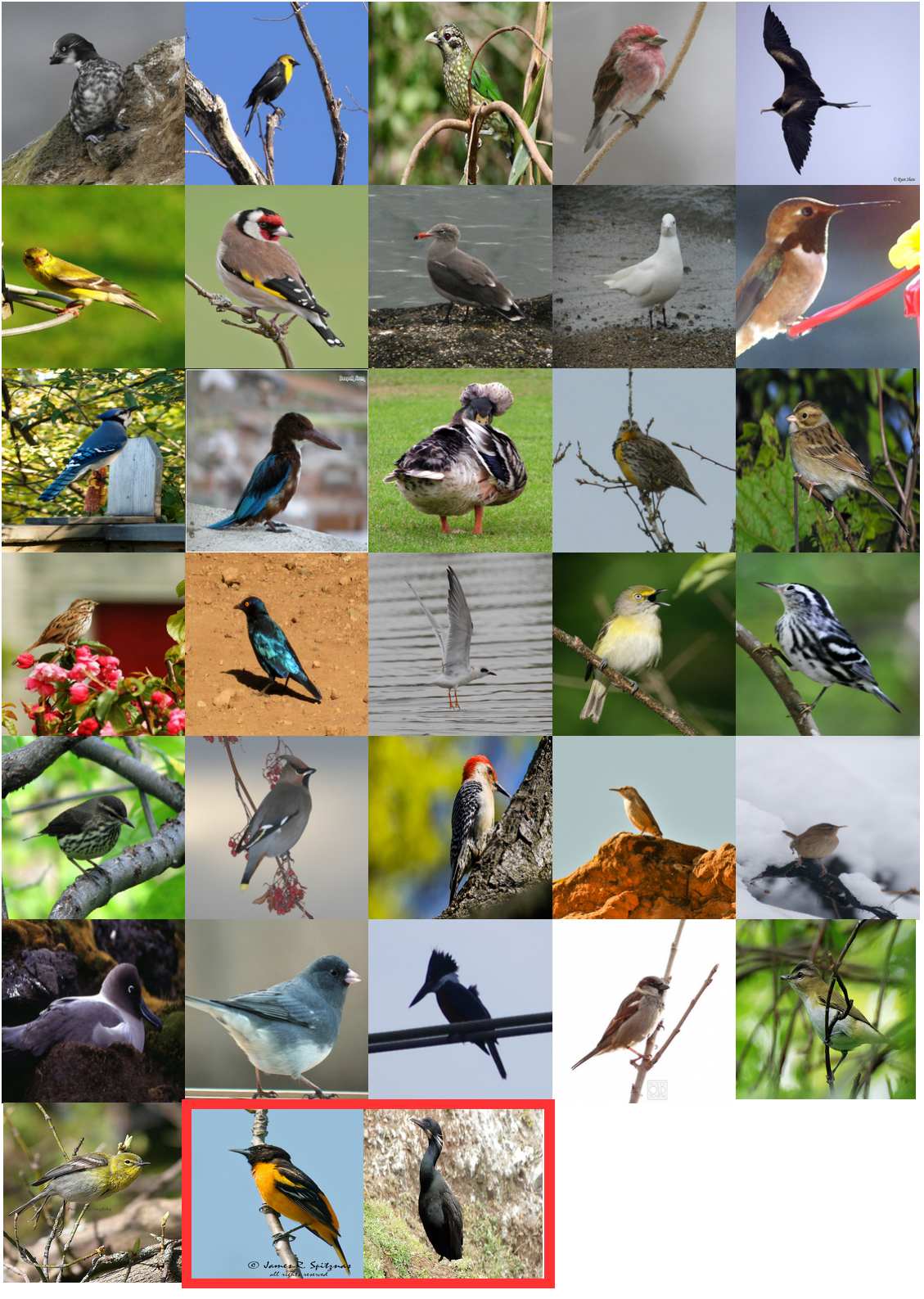}
		\caption{After iteration 4}
		\label{fig:seeds4}
	\end{subfigure}
	
	\caption{New classes (in red boxes) added to the initial seed set (Fig.~\ref{fig:seeds0}) after the first four iterations of VSM. Corresponding results in the semantic space can be found in Fig.~\ref{fig:semantic_vsm}. Visual diversity can be observed as representatives of various families like {\it sparrows}, {\it albatrosses}, {\it cormorants}, {\it kingfishers} and others have been acquired by VSM. Rare attributes like {\it purple underparts}, {\it needle-shaped bill} and others have also been captured within these classes}
	\Description{Capturing diversity and rarity every iteration}
	\label{fig:vsm_seen}
\end{figure}

Figure~\ref{fig:vsm_candidates} shows the $t$ samples for the first four VSM iterations considered visually most diverse from the existing seed classes at the start of every iteration. For CUB data set, $t = 13$ (Sec.~\ref{sec:5.2}). New classes added to the initial seed set at every iteration have been shown in Fig.~\ref{fig:vsm_seen}, where classes at the end of iteration $i$ serve as the seed classes at the start of iteration $i+1$ (for $i = 1,2,3$). The initial 25 seed classes (Fig.~\ref{fig:seeds0}) are used for acquiring new classes in iteration 1 of VSM. 

For iteration 1, Fig.~\ref{fig:vsm1} presents different kinds of {\it swallows, albatrosses} and {\it sparrows} which are different from the birds captured in the seed set. At the start of iteration 1, the top-5 rare and common attributes captured from the existing seed classes are shown in Tab.~\ref{tab:vsm_rare_common}. These lists are obtained according to the fraction of seed class images that the attributes appear in, and hence can approximately be verified from the visual images of the seed classes (Fig.~\ref{fig:seeds0}). The added classes after iteration 1 are {\it Sooty albatross} and {\it Dark-eyed junco}. Referring to the semantic vectors for these two classes, we find that both of them marginally exhibit the top-5 rare attributes, except {\it green leg}. Moreover, as expected, both the added classes exhibit huge amounts of some of the top-5 common attributes like {\it black eye} and {\it solid belly pattern}. After adding these new classes to the previous seed set, the list of top-5 rare attributes changes in the second iteration of VSM, indicating that the seed set has now been enriched with rare attributes. The list of top-5 common attributes remains mostly the same, as these attributes are already the most abundant ones in the object domain.

\begin{table}[t]
	\begin{center}
		\caption{The top five rare and common attributes captured by analyzing the semantic space of the classes in the seed set at the start of the first four iterations of VSM. For each iteration, the attributes are shown in descending order of weights assigned to them, computed using Eqs.~\ref{eq:6},~\ref{eq:13}, and~\ref{eq:4}}
		\label{tab:vsm_rare_common}
		\begin{tabular}{lll}
			\hline
			\multirow{1}{*}{{\bf Iteration}}   & 
			\multirow{1}{*}{{\bf Top Rare}}   & 
			\multirow{1}{*}{{\bf Top Common}}  \\
			
			
			\hline
			
			\multirow{5}{*}{1}
			&Purple under-tail &Small size \\
			&Primarily purple &Solid belly pattern \\
			&Purple nape  &Bill shorter than head\\
			&Green leg &Solid breast pattern \\
			&Purple underparts &Black eye \\
			\hline
			\multirow{5}{*}{2}
			&Pink eye &Small size \\
			&Purple breast &Bill shorter than head \\
			&Green leg &Solid breast pattern\\
			&Purple under-tail &Solid belly pattern \\
			&Green bill &Black eye \\
			\hline
			
			\multirow{5}{*}{3}
			&Pink eye &Small size \\
			&Purple eye &Bill shorter than head  \\
			&Green leg &Solid breast pattern \\
			&Purple under-tail &Solid belly pattern \\
			&Pink under-tail &Black eye\\
			
			\hline
			
			\multirow{5}{*}{4}
			&Owl-like shape &Small size \\
			&Pink eye &Bill shorter than head  \\
			&Purple eye &Solid breast pattern \\
			&Purple under-tail &Solid belly pattern \\
			&Green leg &Black eye\\

			\hline
		\end{tabular}
		
	\end{center}
\end{table}

\subsection{Parameter sensitivity} \label{sec:6.3}
During VSM, attribute-weights are computed based on the semantics of classes in $\mathcal{Z}$ only (Eq.~\ref{eq:4}). Diversity and rarity expressed by such a small portion of the object domain should not dictate the selection of too many classes at a time. Figure~\ref{fig:hyperparam} suggests that $q$ is not very sensitive to ZSL model performance, so we set low values of $q$ for VSM to steadily explore the object domain while expanding the set $\mathcal{Z}$. Since the average image count per class for SUN is relatively lower than CUB, we set $q$ to be higher for SUN to train the feature extractor effectively.

\begin{figure}
	\centering
	\begin{subfigure}{0.34\textwidth}
		\includegraphics[scale=0.111]{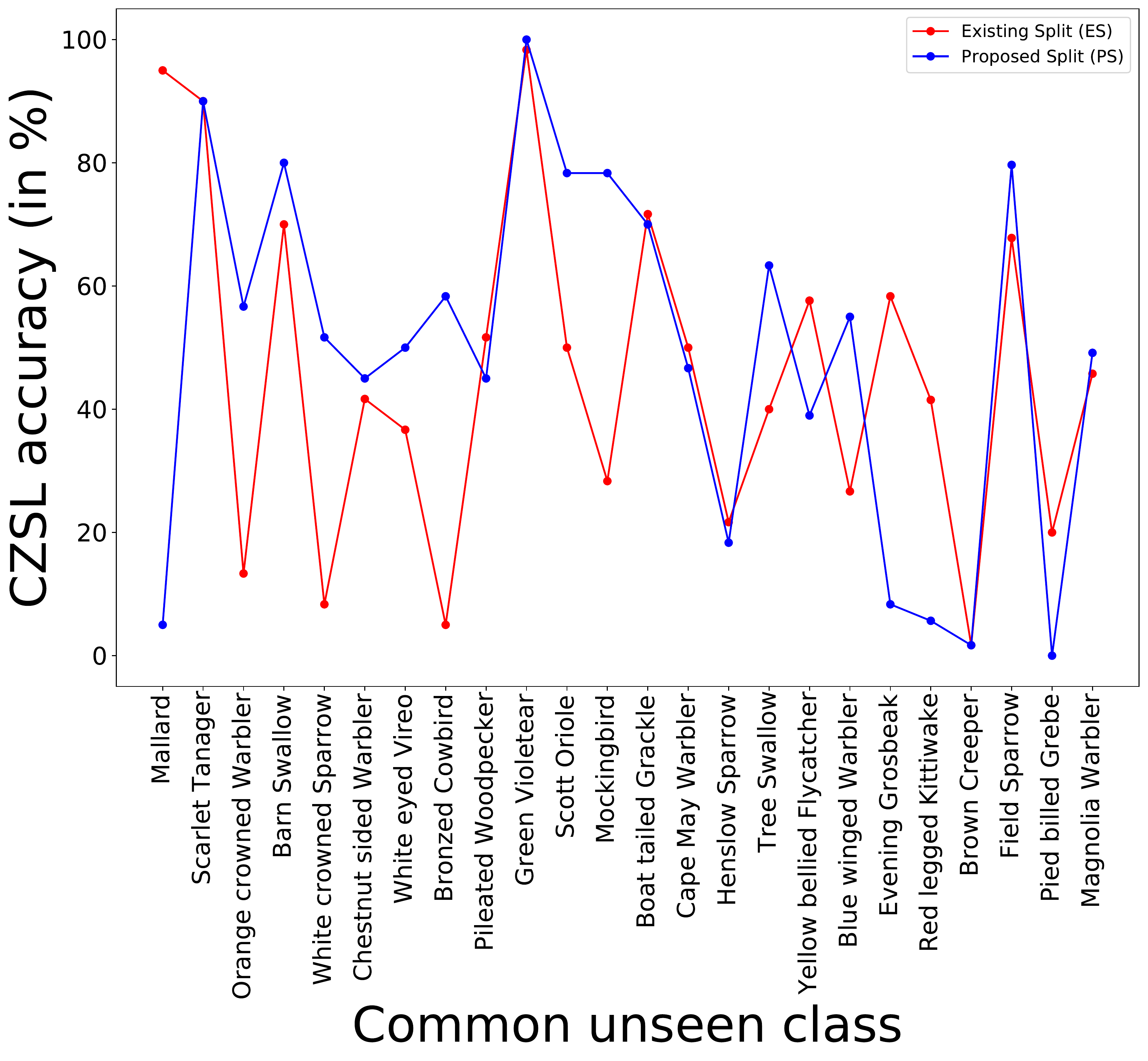}
		\caption{CUB, DeViSE}
		\label{fig:adcub}
	\end{subfigure}
	\hfil
	\begin{subfigure}{0.34\textwidth}
		\includegraphics[scale=0.111]{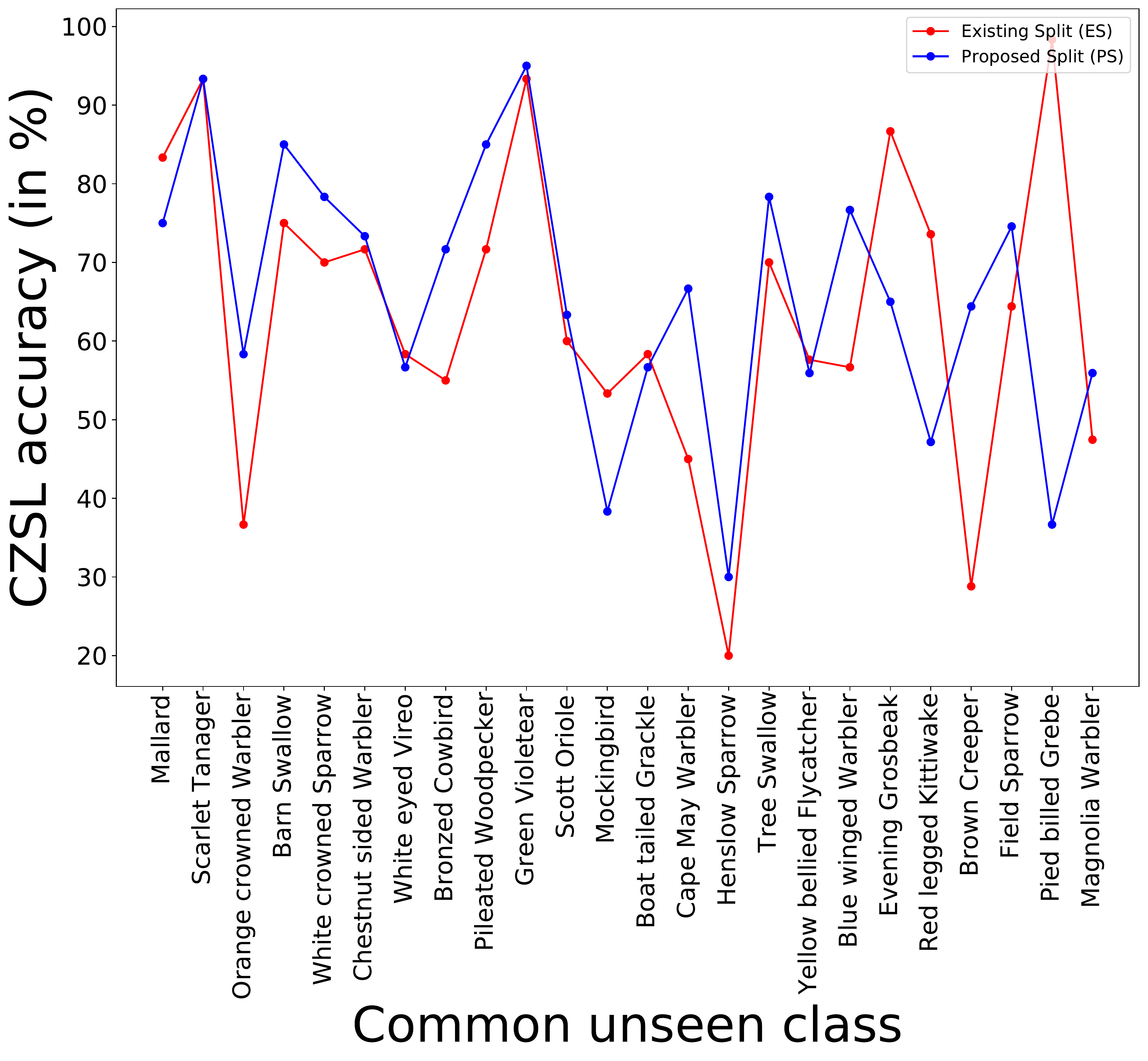}
		\caption{CUB, TF-VAEGAN}
		\label{fig:atcub}
	\end{subfigure}
	\hfil
	\begin{subfigure}{0.34\textwidth}
		\includegraphics[scale=0.111]{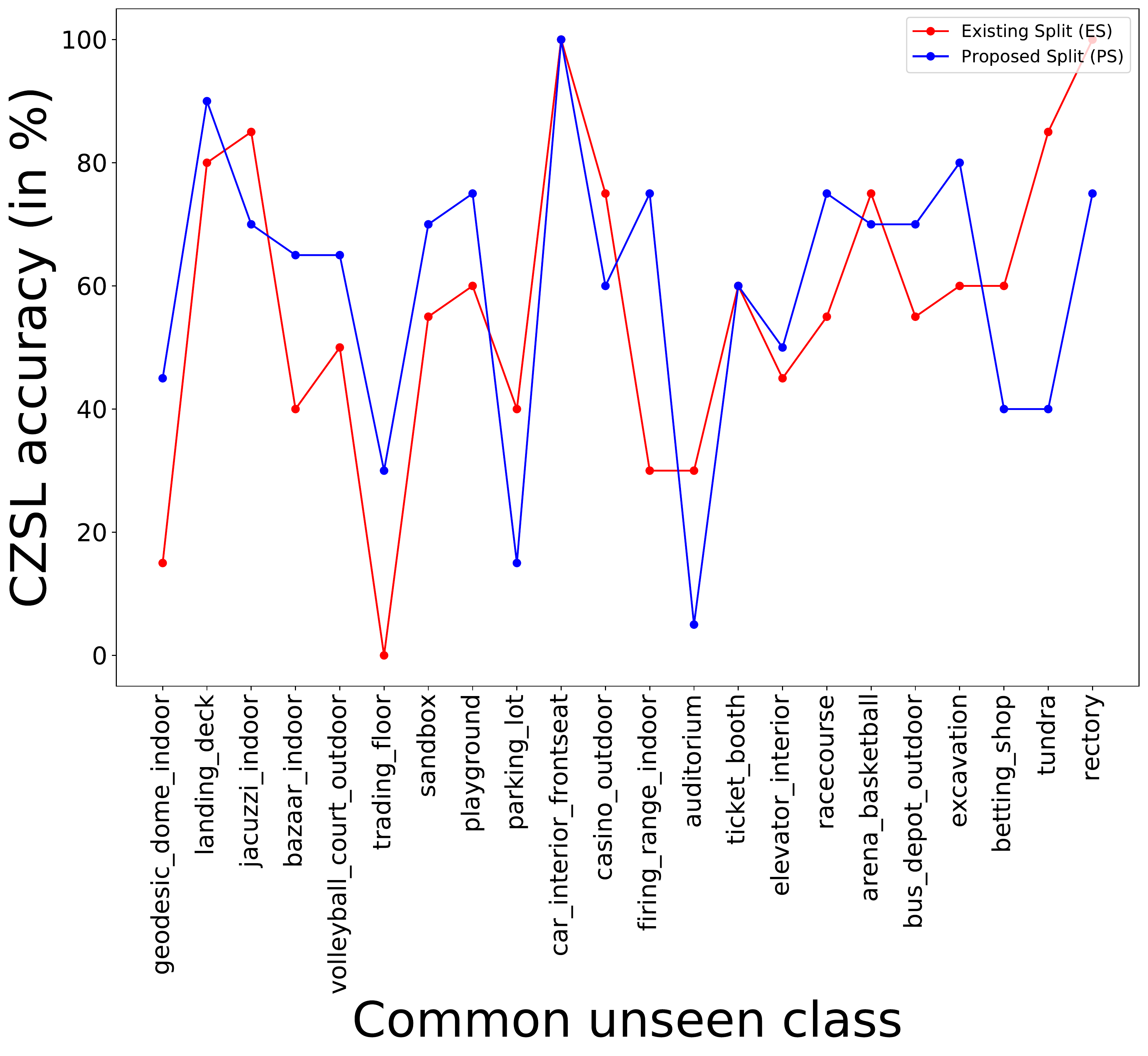}
		\caption{SUN, DeViSE}
		\label{fig:adsun}
	\end{subfigure}
	\hfil
	\begin{subfigure}{0.34\textwidth}
		\includegraphics[scale=0.111]{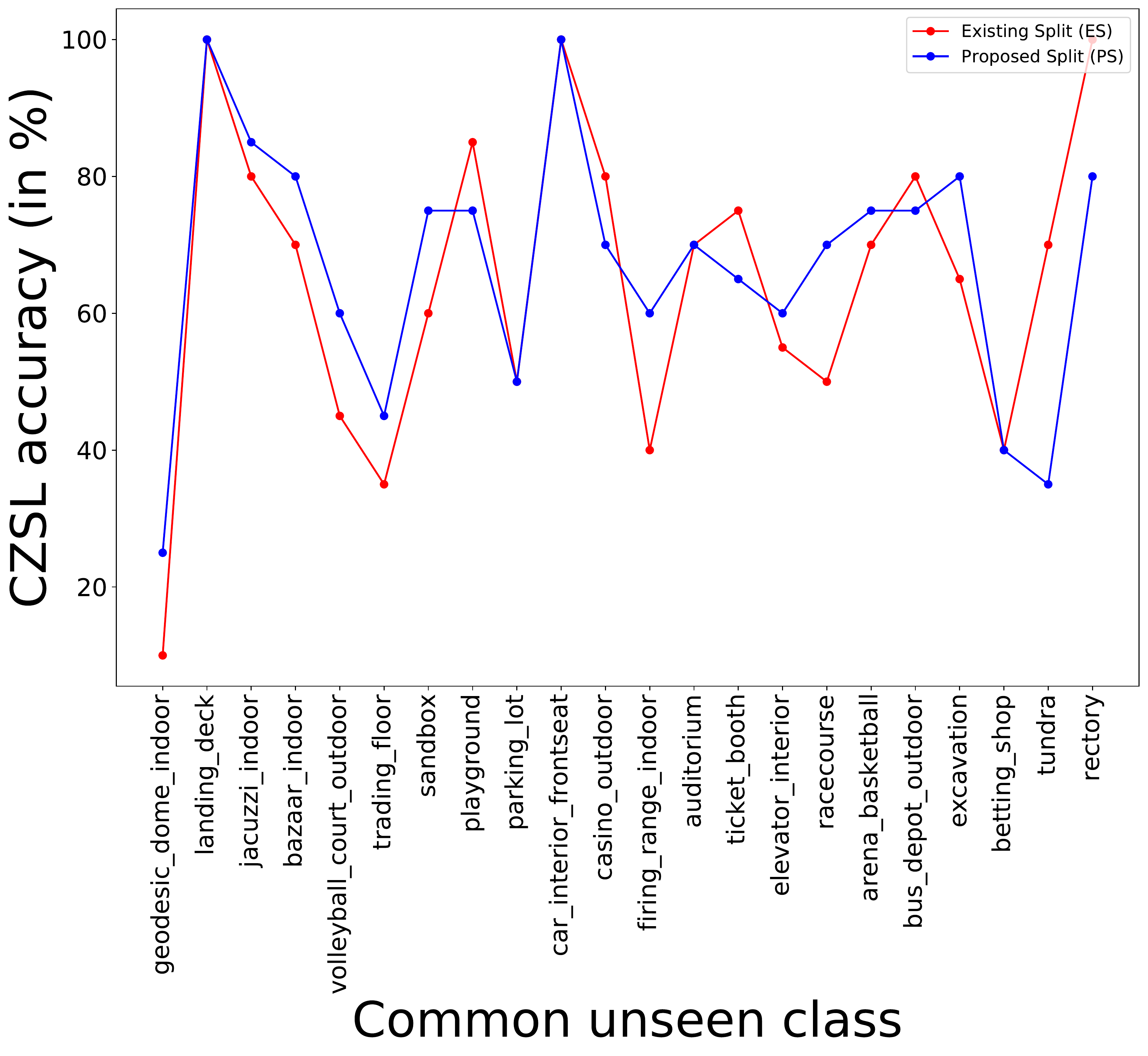}
		\caption{SUN, TF-VAEGAN}
		\label{fig:atsun}
	\end{subfigure}
	
	\begin{subfigure}{0.34\textwidth}
		\includegraphics[scale=0.111]{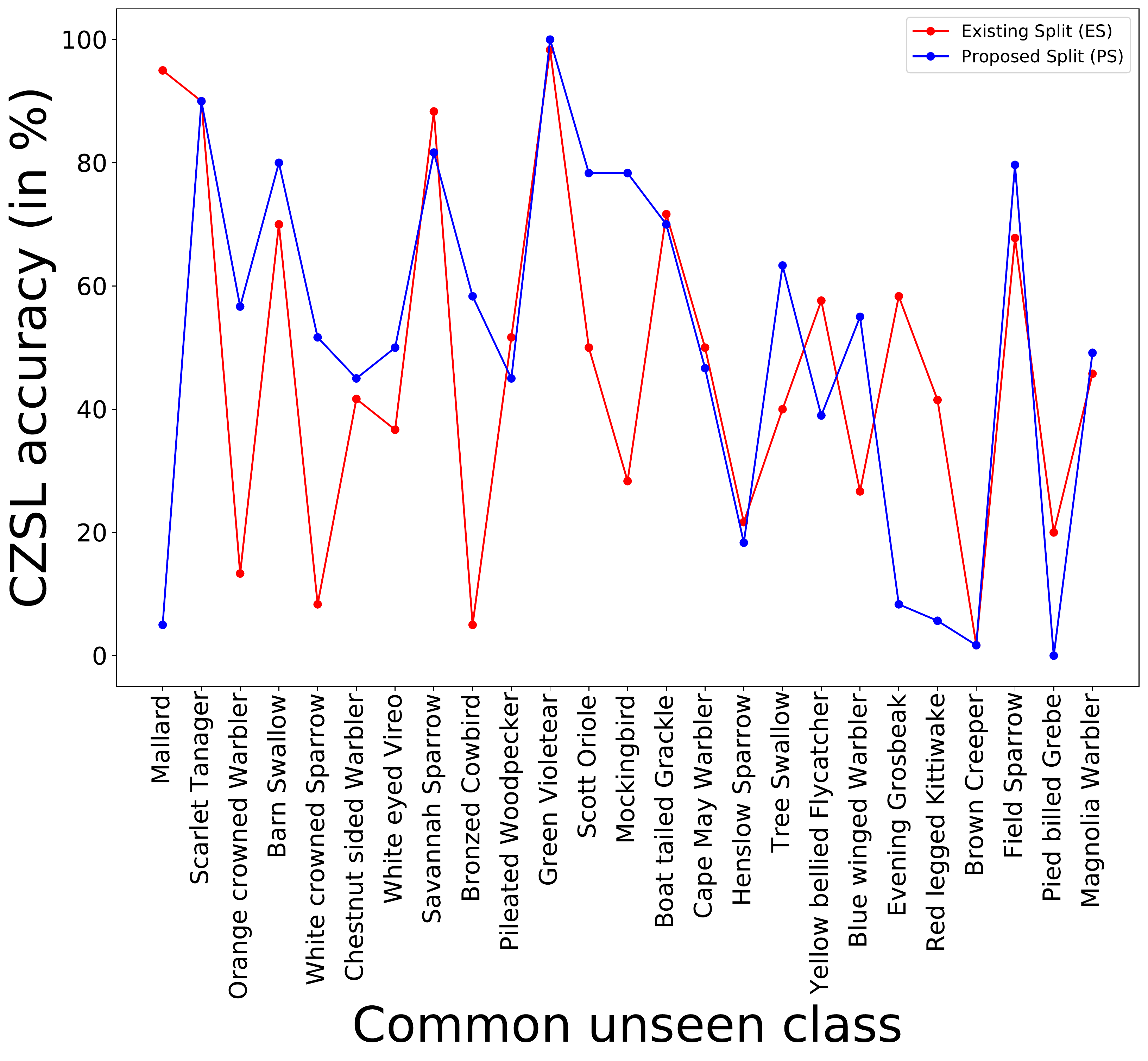}
		\caption{CUB, DeViSE}
		\label{fig:acdsub}
	\end{subfigure}
	\hfil
	\begin{subfigure}{0.34\textwidth}
		\includegraphics[scale=0.111]{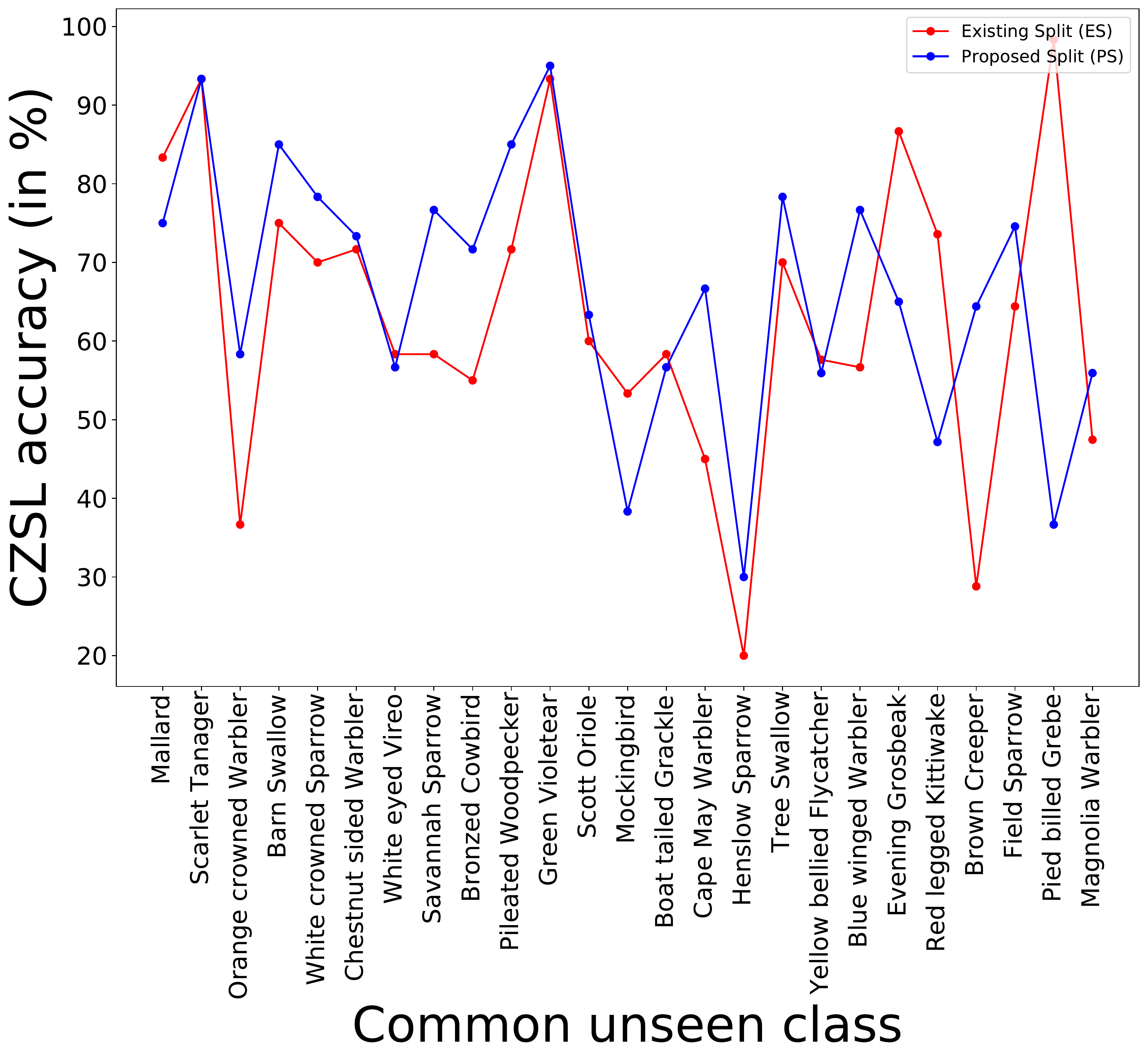}
		\caption{CUB, TF-VAEGAN}
		\label{fig:actcub}
	\end{subfigure}
	\hfil
	\begin{subfigure}{0.34\textwidth}
		\includegraphics[scale=0.111]{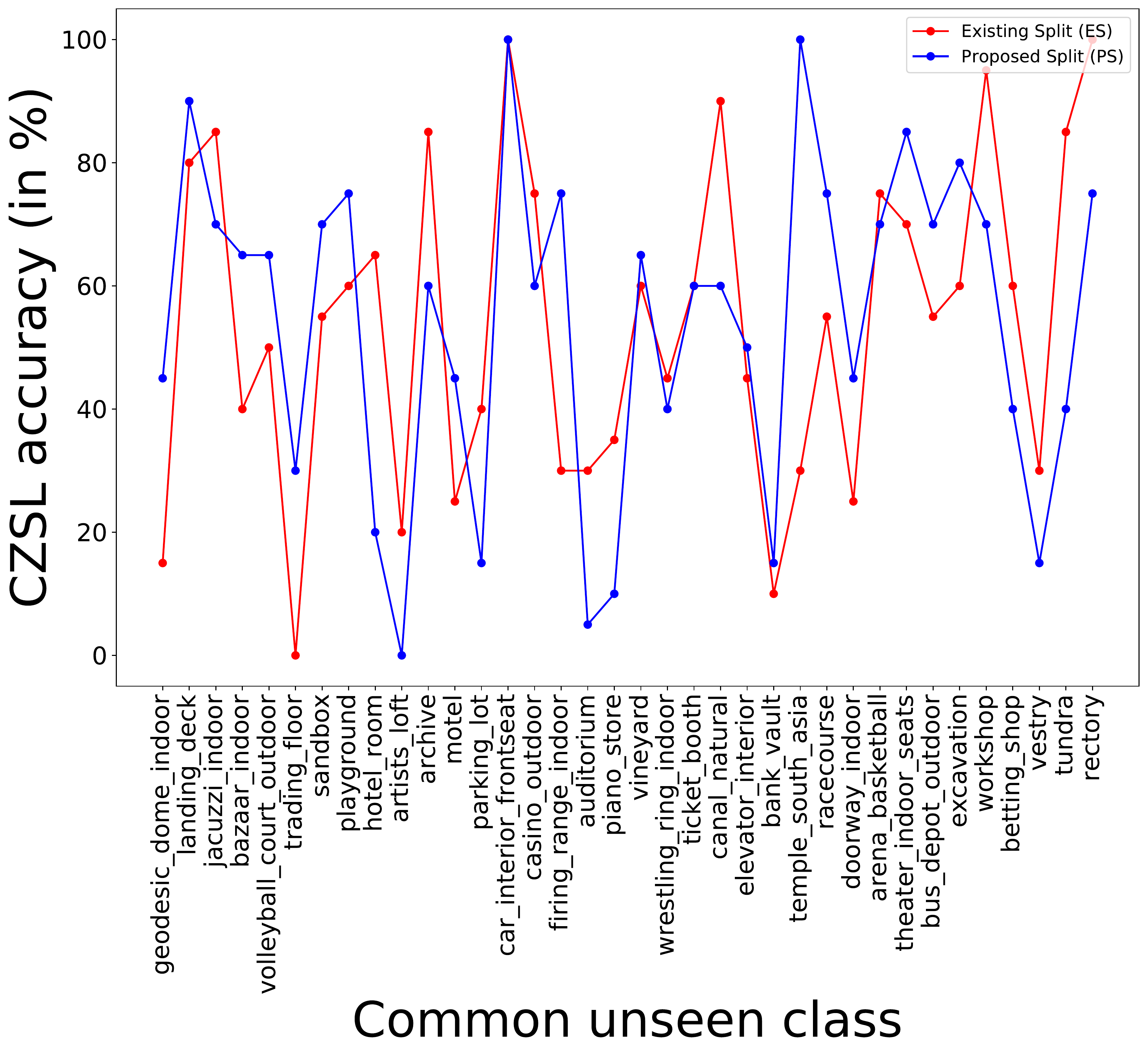}
		\caption{SUN, TF-VAEGAN}
		\label{fig:acdsun}
	\end{subfigure}
	\hfil
	\begin{subfigure}{0.34\textwidth}
		\includegraphics[scale=0.111]{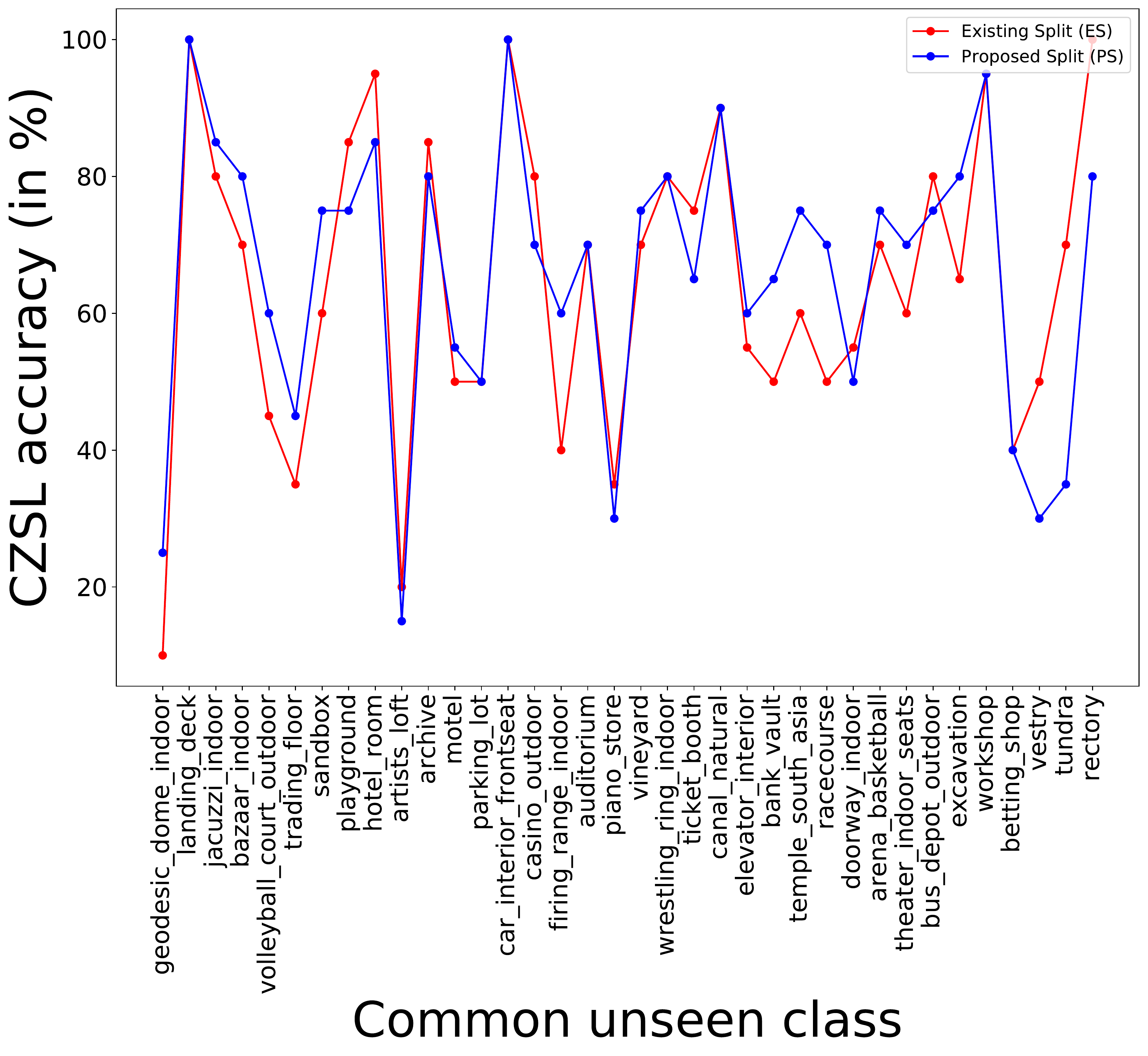}
		\caption{SUN, TF-VAEGAN}
		\label{fig:actsun}
	\end{subfigure}

	\caption{Class-wise accuracy (in \%) in the CZSL setting for test classes from $\mathcal{U}_{com}^1$ (obtained after randomly splitting set $\mathcal{U}$ of Existing Split (ES)), containing at least one rare attribute ((a)--(d)) or one common attribute ((e)--(h)). The curves are obtained by evaluating the performance of two trained models -- DeViSE~\cite{frome2013devise} and TF-VAEGAN~\cite{narayan2020latent}. Common unseen classes are the test classes on which we evaluate models trained with seen classes from Existing Splits (ES) and Proposed Splits (PS)}
	\Description{Impact on class-wise test accuracy of ZSL models by selecting suitable seen classes for ZSL model training}
	\label{fig:classwise}
\end{figure}

\begin{table}[t]
	\begin{center}
		\caption{Distribution of rare and common attributes for different random splits of $\mathcal{U}\textsubscript{E}$. $A = $ total number of attributes; $N_{\Tilde{C}} = N_{s} + N_{\Tilde{u}}$; A\textsubscript{R} and A\textsubscript{C} are the number of rare and common attributes; Y\textsubscript{R} and Y\textsubscript{C} are the number of common unseen classes having at least one rare and one common attribute respectively}
		\label{tab:3}
		\begin{tabular}{lcccc}
			\hline
			\multirow{1}{*}{{\bf Dataset}}   & 
			\multirow{1}{*}{\bf A / N\textsubscript{$\Tilde{C}$} / N\textsubscript{u\textsubscript{com}}}   & 
			\multirow{1}{*}{\textbf{Split}}  & 
			\multirow{1}{*}{\bf A\textsubscript{R} / A\textsubscript{C}}  & 
			\multirow{1}{*}{\bf Y\textsubscript{R} / Y\textsubscript{C}}  \\
			
			\hline
			
			\multirow{3}{*}{CUB} &
			\multirow{3}{*}{312 / 175 / 25}
			&1 &24 / 9 &24 / 25  \\
			& &2 &22 / 9 &22 / 25  \\
			& &3 &22 / 10 &19 / 25  \\
			
			\hline
			
			\multirow{3}{*}{SUN} &
			\multirow{3}{*}{102 / 681 / 36} 
			&1 &7 / 2 &22 / 36  \\
			& &2 &7 / 2 &19 / 36  \\
			& &3 &6 / 2 &15 / 36  \\
			
			\hline
		\end{tabular}
	\end{center}
\end{table}

\begin{table}[t]
	\begin{center}
		\caption{Understanding object domain via attributes. Five rare and common attributes (designated as described in Sec.~\ref{sec:6.4}) are listed for the three different object domains for which we show our results. {$\Tilde{\mathcal{C}^X}$} denotes the classes in the object domain acquired as $\Tilde{\mathcal{C}^X} = \mathcal{S}\textsubscript{E} \, \cup \, (\mathcal{U}\textsubscript{E} \setminus \mathcal{U}_{com}^X$). Here, $\mathcal{U}_{com}^{X}$ denotes the set of common unseen classes for both ES and PS separated out for fair evaluation, created by $X^{th}$ random split of set $\mathcal{U}\textsubscript{E}$.  (+$p$) indicates set contains $p$ more attributes}
		\label{tab:rare_common}
		\begin{tabular}{lll}
			\hline
			\multirow{1}{*}{{\bf Domain}}   & 
			\multirow{1}{*}{{\bf Rare}}   & 
			\multirow{1}{*}{{\bf Common}}  \\
			
			
			\hline
			
			\multirow{5}{*}{$\Tilde{\mathcal{C}^1}$} 
			&Needle-shaped bill &Black bill \\
			&Pink throat &Notched tail \\
			&Red back &Small size \\
			&Purple eye &Rounded wings \\
			&Owl-like shape (+19) &Black eye (+4) \\
			\hline
			
			\multirow{5}{*}{$\Tilde{\mathcal{C}^2}$} 
			&Needle-shaped bill &Rounded wing \\
			&Purple underparts &Solid breast pattern \\
			&Pink forehead &Notched tail \\
			&Green leg &Bill shorter than head \\
			&Orange eye (+17) &Solid belly pattern (+4) \\
			\hline
			
			\multirow{5}{*}{$\Tilde{\mathcal{C}^3}$}
			&Red upper-tail &Rounded wing \\
			&Pink crown &Small size \\
			&Green crown &Solid back pattern \\
			&Purple breast &Black eye \\
			&Red underparts (+17) &Black bill (+5) \\

			\hline
		\end{tabular}
		
	\end{center}
\end{table}

\subsection{Acknowledging rarity in the object domain}
\label{sec:6.4}
For attribute-based data, an object class is uniquely characterized by its attributes, so it can be reasoned that the more rare attributes a class exhibits, the higher its probability of being a rare class. To test the semantic knowledge gained by our framework about the object domain, we develop a notion for designating attributes as either {\it rare} or {\it common} using semantic information from $\mathcal{P}(\Tilde{\mathcal{C}})$. Sets $IA(\Tilde{\mathcal{C}})$ and $UA(\Tilde{\mathcal{C}})$ are developed using Eqs.~\ref{eq:9} and~\ref{eq:10} and their member attributes are discarded. Then, we analyze $\mathcal{B}({\Tilde{C}})$ (obtained using Eq.~\ref{eq:11}) and designate the attributes which appear in less than 5\% of all classes in $\Tilde{\mathcal{C}}$ as rare, and those appearing in more than 50\% of the classes as common attributes. Table~\ref{tab:3} indicates that there are fewer rare attributes in SUN as compared to CUB. This was expected as the attributes in SUN are observed in many different contexts~\cite{patterson2014sun} and hence appear for many classes. On the other hand, several attributes in CUB are visual variants of a single, broader attribute~\cite{WahCUB_200_2011}. Hence, several of these attributes are exhibited by a few classes only.

We report a few rare and common attributes for each object domain (i.e. $\Tilde{\mathcal{C}^1}, \Tilde{\mathcal{C}^2}$ and $\Tilde{\mathcal{C}^3}$) in Tab.~\ref{tab:rare_common}. Thereafter, looking back at the seed classes acquired from $\Tilde{\mathcal{C}^2}$ at the end of stage 1 (Fig.~\ref{fig:seeds0}), we find that our seed-set construction process indeed picks an initial seed set which is not only diverse enough, but also captures the rarity from the semantic space of the object domain. For example, 
Fig.~\ref{fig:seeds0} shows birds exhibiting {\it needle-shaped bill} and {\it orange eye} (in colored boxes), which are rare attributes considering the domain $\Tilde{\mathcal{C}^2}$ (see Fig.~\ref{tab:rare_common}).

Table~\ref{tab:2} conveys that training ZSL models with seen classes that capture rarity in the object domain well enough enhance the models' capability to recognize novel classes exhibiting rare attributes. The class-wise top-1 accuracy after training with two ZSL models --- DeViSE~\cite{frome2013devise} (a compatibility learning framework) and TF-VAEGAN~\cite{narayan2020latent} (a generative model-based framework) --- is depicted in Fig.~\ref{fig:classwise} for test classes from $\mathcal{U}_{com}^1$ exhibiting at least one rare or common attribute. It is evident that models in $\Phi_P$ (those trained by seen classes selected by DiRaC-I) recognize novel classes more accurately.

\section{Conclusion} \label{disc}
In this paper, we propose a novel framework called DiRaC-I for identifying the most suitable classes from the available database that can be used to train zero-shot models. Specifically, we emphasize capturing both visual diversity and semantic rarity of an object domain through our framework, inspired by Active Learning. Extensive experiments on two challenging fine-grained data sets verified that zero-shot models trained with classes acquired by DiRaC-I perform better than models trained with predetermined classes. We limit our work to these data sets for fair comparison as they have a balanced image count across all the classes, unlike certain others like AwA2~\cite{xian2018zero}. This ensures that even if seen classes for ES and PS are different, it does not adversely affect ZSL model performance just due to a huge difference in number of training images. Additionally, we work only with human-annotated attributes to account for rarity as they are more semantically descriptive and interpretable than word vector representations of classes. Such an attribute space is consistent with our real-life goal of zero-shot methods working in a specific object domain. However, manually defining attribute ontology is expensive. Hence, an extension of DiRaC-I that can work with word vector spaces is worth investigating.

\bibliographystyle{ACM-Reference-Format}
\bibliography{dirac-i-ref}


\begin{thebibliography}{61}


\ifx \showCODEN    \undefined \def \showCODEN     #1{\unskip}     \fi
\ifx \showDOI      \undefined \def \showDOI       #1{#1}\fi
\ifx \showISBNx    \undefined \def \showISBNx     #1{\unskip}     \fi
\ifx \showISBNxiii \undefined \def \showISBNxiii  #1{\unskip}     \fi
\ifx \showISSN     \undefined \def \showISSN      #1{\unskip}     \fi
\ifx \showLCCN     \undefined \def \showLCCN      #1{\unskip}     \fi
\ifx \shownote     \undefined \def \shownote      #1{#1}          \fi
\ifx \showarticletitle \undefined \def \showarticletitle #1{#1}   \fi
\ifx \showURL      \undefined \def \showURL       {\relax}        \fi
\providecommand\bibfield[2]{#2}
\providecommand\bibinfo[2]{#2}
\providecommand\natexlab[1]{#1}
\providecommand\showeprint[2][]{arXiv:#2}

\bibitem[Akata et~al\mbox{.}(2016)]%
        {akata2015label}
\bibfield{author}{\bibinfo{person}{Zeynep Akata}, \bibinfo{person}{Florent
  Perronnin}, \bibinfo{person}{Zaid Harchaoui}, {and} \bibinfo{person}{Cordelia
  Schmid}.} \bibinfo{year}{2016}\natexlab{}.
\newblock \showarticletitle{Label-embedding for image classification}.
\newblock \bibinfo{journal}{\emph{IEEE transactions on pattern analysis and
  machine intelligence}} \bibinfo{volume}{38}, \bibinfo{number}{7}
  (\bibinfo{year}{2016}), \bibinfo{pages}{1425--1438}.
\newblock


\bibitem[Akata et~al\mbox{.}(2015)]%
        {akata2015evaluation}
\bibfield{author}{\bibinfo{person}{Zeynep Akata}, \bibinfo{person}{Scott Reed},
  \bibinfo{person}{Daniel Walter}, \bibinfo{person}{Honglak Lee}, {and}
  \bibinfo{person}{Bernt Schiele}.} \bibinfo{year}{2015}\natexlab{}.
\newblock \showarticletitle{Evaluation of output embeddings for fine-grained
  image classification}. In \bibinfo{booktitle}{\emph{CVPR}}.
  \bibinfo{pages}{2927--2936}.
\newblock


\bibitem[Bansal et~al\mbox{.}(2018)]%
        {bansal2018zero}
\bibfield{author}{\bibinfo{person}{Ankan Bansal}, \bibinfo{person}{Karan
  Sikka}, \bibinfo{person}{Gaurav Sharma}, \bibinfo{person}{Rama Chellappa},
  {and} \bibinfo{person}{Ajay Divakaran}.} \bibinfo{year}{2018}\natexlab{}.
\newblock \showarticletitle{Zero-shot object detection}. In
  \bibinfo{booktitle}{\emph{Proceedings of the European Conference on Computer
  Vision (ECCV)}}. \bibinfo{pages}{384--400}.
\newblock


\bibitem[Bendale and Boult(2016)]%
        {bendale2016towards}
\bibfield{author}{\bibinfo{person}{Abhijit Bendale} {and}
  \bibinfo{person}{Terrance~E Boult}.} \bibinfo{year}{2016}\natexlab{}.
\newblock \showarticletitle{Towards open set deep networks}. In
  \bibinfo{booktitle}{\emph{CVPR}}. \bibinfo{pages}{1563--1572}.
\newblock


\bibitem[Cao and Zhang(2022)]%
        {cao2022learning}
\bibfield{author}{\bibinfo{person}{Congqi Cao} {and} \bibinfo{person}{Yanning
  Zhang}.} \bibinfo{year}{2022}\natexlab{}.
\newblock \showarticletitle{Learning to compare relation: Semantic alignment
  for few-shot learning}.
\newblock \bibinfo{journal}{\emph{IEEE Transactions on Image Processing}}
  \bibinfo{volume}{31} (\bibinfo{year}{2022}), \bibinfo{pages}{1462--1474}.
\newblock


\bibitem[Changpinyo et~al\mbox{.}(2016)]%
        {changpinyo2016synthesized}
\bibfield{author}{\bibinfo{person}{Soravit Changpinyo},
  \bibinfo{person}{Wei-Lun Chao}, \bibinfo{person}{Boqing Gong}, {and}
  \bibinfo{person}{Fei Sha}.} \bibinfo{year}{2016}\natexlab{}.
\newblock \showarticletitle{Synthesized classifiers for zero-shot learning}. In
  \bibinfo{booktitle}{\emph{CVPR}}. \bibinfo{pages}{5327--5336}.
\newblock


\bibitem[Chen et~al\mbox{.}(2021)]%
        {chen2021deep}
\bibfield{author}{\bibinfo{person}{Bingzhi Chen}, \bibinfo{person}{Yishu Liu},
  \bibinfo{person}{Zheng Zhang}, \bibinfo{person}{Yingjian Li},
  \bibinfo{person}{Zhao Zhang}, \bibinfo{person}{Guangming Lu}, {and}
  \bibinfo{person}{Hongbing Yu}.} \bibinfo{year}{2021}\natexlab{}.
\newblock \showarticletitle{Deep Active Context Estimation for Automated
  COVID-19 Diagnosis}.
\newblock \bibinfo{journal}{\emph{ACM Transactions on Multimedia Computing,
  Communications, and Applications (TOMM)}} \bibinfo{volume}{17},
  \bibinfo{number}{3s} (\bibinfo{year}{2021}), \bibinfo{pages}{1--22}.
\newblock


\bibitem[Chen and Huang(2021)]%
        {chen2021elaborative}
\bibfield{author}{\bibinfo{person}{Shizhe Chen} {and} \bibinfo{person}{Dong
  Huang}.} \bibinfo{year}{2021}\natexlab{}.
\newblock \showarticletitle{Elaborative rehearsal for zero-shot action
  recognition}. In \bibinfo{booktitle}{\emph{Proceedings of the IEEE/CVF
  International Conference on Computer Vision}}. \bibinfo{pages}{13638--13647}.
\newblock


\bibitem[Dligach and Palmer(2011)]%
        {dligach-palmer-2011-good}
\bibfield{author}{\bibinfo{person}{Dmitriy Dligach} {and}
  \bibinfo{person}{Martha Palmer}.} \bibinfo{year}{2011}\natexlab{}.
\newblock \showarticletitle{Good Seed Makes a Good Crop: {A}ccelerating Active
  Learning Using Language Modeling}. In \bibinfo{booktitle}{\emph{Proceedings
  of the 49th Annual Meeting of the Association for Computational Linguistics:
  Human Language Technologies}}. \bibinfo{pages}{6--10}.
\newblock


\bibitem[Fei-Fei et~al\mbox{.}(2006)]%
        {fei2006one}
\bibfield{author}{\bibinfo{person}{Li Fei-Fei}, \bibinfo{person}{Robert
  Fergus}, {and} \bibinfo{person}{Pietro Perona}.}
  \bibinfo{year}{2006}\natexlab{}.
\newblock \showarticletitle{One-shot learning of object categories}.
\newblock \bibinfo{journal}{\emph{IEEE transactions on pattern analysis and
  machine intelligence}} \bibinfo{volume}{28}, \bibinfo{number}{4}
  (\bibinfo{year}{2006}), \bibinfo{pages}{594--611}.
\newblock


\bibitem[Felix et~al\mbox{.}(2018)]%
        {felix2018multi}
\bibfield{author}{\bibinfo{person}{Rafael Felix}, \bibinfo{person}{Vijay B.~G.
  Kumar}, \bibinfo{person}{Ian Reid}, {and} \bibinfo{person}{Gustavo
  Carneiro}.} \bibinfo{year}{2018}\natexlab{}.
\newblock \showarticletitle{Multi-modal cycle-consistent generalized zero-shot
  learning}. In \bibinfo{booktitle}{\emph{ECCV}}. \bibinfo{pages}{21--37}.
\newblock


\bibitem[Feng and Zhao(2021)]%
        {feng2020transfer}
\bibfield{author}{\bibinfo{person}{Liangjun Feng} {and}
  \bibinfo{person}{Chunhui Zhao}.} \bibinfo{year}{2021}\natexlab{}.
\newblock \showarticletitle{Transfer increment for generalized zero-shot
  learning}.
\newblock \bibinfo{journal}{\emph{{IEEE} Transactions on Neural Networks and
  Learning Systems}} \bibinfo{volume}{32}, \bibinfo{number}{6}
  (\bibinfo{year}{2021}), \bibinfo{pages}{2506--2520}.
\newblock


\bibitem[Frome et~al\mbox{.}(2013)]%
        {frome2013devise}
\bibfield{author}{\bibinfo{person}{Andrea Frome}, \bibinfo{person}{Greg~S.
  Corrado}, \bibinfo{person}{Jon Shlens}, \bibinfo{person}{Samy Bengio},
  \bibinfo{person}{Jeff Dean}, \bibinfo{person}{Marc'Aurelio Ranzato}, {and}
  \bibinfo{person}{Tomas Mikolov}.} \bibinfo{year}{2013}\natexlab{}.
\newblock \showarticletitle{{DeViSE}: {A} deep visual-semantic embedding
  model}. In \bibinfo{booktitle}{\emph{NIPS}}. \bibinfo{pages}{2121--2129}.
\newblock


\bibitem[Hanneke(2014)]%
        {hanneke2014theory}
\bibfield{author}{\bibinfo{person}{Steve Hanneke}.}
  \bibinfo{year}{2014}\natexlab{}.
\newblock \bibinfo{booktitle}{\emph{Theory of disagreement-based active
  learning}}.
\newblock \bibinfo{publisher}{Now Foundations and Trends}.
\newblock


\bibitem[He et~al\mbox{.}(2016)]%
        {he2016deep}
\bibfield{author}{\bibinfo{person}{Kaiming He}, \bibinfo{person}{Xiangyu
  Zhang}, \bibinfo{person}{Shaoqing Ren}, {and} \bibinfo{person}{Jian Sun}.}
  \bibinfo{year}{2016}\natexlab{}.
\newblock \showarticletitle{Deep residual learning for image recognition}. In
  \bibinfo{booktitle}{\emph{CVPR}}. \bibinfo{pages}{770--778}.
\newblock


\bibitem[Howard et~al\mbox{.}(2017)]%
        {howard2017mobilenets}
\bibfield{author}{\bibinfo{person}{Andrew~G Howard}, \bibinfo{person}{Menglong
  Zhu}, \bibinfo{person}{Bo Chen}, \bibinfo{person}{Dmitry Kalenichenko},
  \bibinfo{person}{Weijun Wang}, \bibinfo{person}{Tobias Weyand},
  \bibinfo{person}{Marco Andreetto}, {and} \bibinfo{person}{Hartwig Adam}.}
  \bibinfo{year}{2017}\natexlab{}.
\newblock \showarticletitle{Mobilenets: Efficient convolutional neural networks
  for mobile vision applications}.
\newblock \bibinfo{journal}{\emph{arXiv preprint arXiv:1704.04861}}
  (\bibinfo{year}{2017}).
\newblock


\bibitem[Huang et~al\mbox{.}(2017)]%
        {huang2017densely}
\bibfield{author}{\bibinfo{person}{Gao Huang}, \bibinfo{person}{Zhuang Liu},
  \bibinfo{person}{Laurens Van Der~Maaten}, {and} \bibinfo{person}{Kilian~Q
  Weinberger}.} \bibinfo{year}{2017}\natexlab{}.
\newblock \showarticletitle{Densely connected convolutional networks}. In
  \bibinfo{booktitle}{\emph{CVPR}}. \bibinfo{pages}{4700--4708}.
\newblock


\bibitem[Ishihara et~al\mbox{.}(2021)]%
        {ishihara2021multi}
\bibfield{author}{\bibinfo{person}{Keishi Ishihara}, \bibinfo{person}{Anssi
  Kanervisto}, \bibinfo{person}{Jun Miura}, {and} \bibinfo{person}{Ville
  Hautamaki}.} \bibinfo{year}{2021}\natexlab{}.
\newblock \showarticletitle{Multi-task Learning with Attention for End-to-end
  Autonomous Driving}. In \bibinfo{booktitle}{\emph{CVPR}}.
  \bibinfo{pages}{2902--2911}.
\newblock


\bibitem[Jiang et~al\mbox{.}(2020)]%
        {jiang2020few}
\bibfield{author}{\bibinfo{person}{Shuqiang Jiang}, \bibinfo{person}{Weiqing
  Min}, \bibinfo{person}{Yongqiang Lyu}, {and} \bibinfo{person}{Linhu Liu}.}
  \bibinfo{year}{2020}\natexlab{}.
\newblock \showarticletitle{Few-shot food recognition via multi-view
  representation learning}.
\newblock \bibinfo{journal}{\emph{ACM Transactions on Multimedia Computing,
  Communications, and Applications (TOMM)}} \bibinfo{volume}{16},
  \bibinfo{number}{3} (\bibinfo{year}{2020}), \bibinfo{pages}{1--20}.
\newblock


\bibitem[Kankuekul et~al\mbox{.}(2012)]%
        {kankuekul2012online}
\bibfield{author}{\bibinfo{person}{Pichai Kankuekul}, \bibinfo{person}{Aram
  Kawewong}, \bibinfo{person}{Sirinart Tangruamsub}, {and}
  \bibinfo{person}{Osamu Hasegawa}.} \bibinfo{year}{2012}\natexlab{}.
\newblock \showarticletitle{Online incremental attribute-based zero-shot
  learning}. In \bibinfo{booktitle}{\emph{CVPR}}. \bibinfo{pages}{3657--3664}.
\newblock


\bibitem[Kaufman and Rousseeuw(2009)]%
        {kaufman2009finding}
\bibfield{author}{\bibinfo{person}{Leonard Kaufman} {and}
  \bibinfo{person}{Peter~J Rousseeuw}.} \bibinfo{year}{2009}\natexlab{}.
\newblock \bibinfo{booktitle}{\emph{Finding groups in data: an introduction to
  cluster analysis}}. Vol.~\bibinfo{volume}{344}.
\newblock \bibinfo{publisher}{John Wiley \& Sons}.
\newblock


\bibitem[Kennedy et~al\mbox{.}(2019)]%
        {kennedy2019unknown}
\bibfield{author}{\bibinfo{person}{Brian~RC Kennedy}, \bibinfo{person}{Kasey
  Cantwell}, \bibinfo{person}{Mashkoor Malik}, \bibinfo{person}{Christopher
  Kelley}, \bibinfo{person}{Jeremy Potter}, \bibinfo{person}{Kelley Elliott},
  \bibinfo{person}{Elizabeth Lobecker}, \bibinfo{person}{Lindsay~McKenna Gray},
  \bibinfo{person}{Derek Sowers}, \bibinfo{person}{Michael~P White},
  {et~al\mbox{.}}} \bibinfo{year}{2019}\natexlab{}.
\newblock \showarticletitle{The unknown and the unexplored: Insights into the
  Pacific deep-sea following NOAA CAPSTONE expeditions}.
\newblock \bibinfo{journal}{\emph{Frontiers in Marine Science}}
  \bibinfo{volume}{6} (\bibinfo{year}{2019}), \bibinfo{pages}{480}.
\newblock


\bibitem[Kodirov et~al\mbox{.}(2017)]%
        {kodirov2017semantic}
\bibfield{author}{\bibinfo{person}{Elyor Kodirov}, \bibinfo{person}{Tao Xiang},
  {and} \bibinfo{person}{Shaogang Gong}.} \bibinfo{year}{2017}\natexlab{}.
\newblock \showarticletitle{Semantic autoencoder for zero-shot learning}. In
  \bibinfo{booktitle}{\emph{CVPR}}. \bibinfo{pages}{4447--4456}.
\newblock


\bibitem[Kunz et~al\mbox{.}(2008)]%
        {kunz2008deep}
\bibfield{author}{\bibinfo{person}{Clayton Kunz}, \bibinfo{person}{Chris
  Murphy}, \bibinfo{person}{Richard Camilli}, \bibinfo{person}{Hanumant Singh},
  \bibinfo{person}{John Bailey}, \bibinfo{person}{Ryan Eustice},
  \bibinfo{person}{Michael Jakuba}, \bibinfo{person}{Ko-ichi Nakamura},
  \bibinfo{person}{Chris Roman}, \bibinfo{person}{Taichi Sato},
  {et~al\mbox{.}}} \bibinfo{year}{2008}\natexlab{}.
\newblock \showarticletitle{Deep sea underwater robotic exploration in the
  ice-covered arctic ocean with AUVs}. In
  \bibinfo{booktitle}{\emph{{IEEE}/{RSJ} International Conference on
  Intelligent Robots and Systems}}. \bibinfo{pages}{3654--3660}.
\newblock


\bibitem[Lampert et~al\mbox{.}(2013)]%
        {lampert2013attribute}
\bibfield{author}{\bibinfo{person}{Christoph~H Lampert},
  \bibinfo{person}{Hannes Nickisch}, {and} \bibinfo{person}{Stefan Harmeling}.}
  \bibinfo{year}{2013}\natexlab{}.
\newblock \showarticletitle{Attribute-based classification for zero-shot visual
  object categorization}.
\newblock \bibinfo{journal}{\emph{PAMI}} \bibinfo{volume}{36},
  \bibinfo{number}{3} (\bibinfo{year}{2013}), \bibinfo{pages}{453--465}.
\newblock


\bibitem[Liu et~al\mbox{.}(2021a)]%
        {liu2021medical}
\bibfield{author}{\bibinfo{person}{Xiangbin Liu}, \bibinfo{person}{Jiesheng
  He}, \bibinfo{person}{Liping Song}, \bibinfo{person}{Shuai Liu}, {and}
  \bibinfo{person}{Gautam Srivastava}.} \bibinfo{year}{2021}\natexlab{a}.
\newblock \showarticletitle{Medical Image Classification based on an Adaptive
  Size Deep Learning Model}.
\newblock \bibinfo{journal}{\emph{ACM Transactions on Multimedia Computing,
  Communications, and Applications (TOMM)}} \bibinfo{volume}{17},
  \bibinfo{number}{3s} (\bibinfo{year}{2021}), \bibinfo{pages}{1--18}.
\newblock


\bibitem[Liu et~al\mbox{.}(2021b)]%
        {liu2021single}
\bibfield{author}{\bibinfo{person}{Xinfang Liu}, \bibinfo{person}{Xiushan Nie},
  \bibinfo{person}{Junya Teng}, \bibinfo{person}{Li Lian}, {and}
  \bibinfo{person}{Yilong Yin}.} \bibinfo{year}{2021}\natexlab{b}.
\newblock \showarticletitle{Single-shot semantic matching network for moment
  localization in videos}.
\newblock \bibinfo{journal}{\emph{ACM Transactions on Multimedia Computing,
  Communications, and Applications (TOMM)}} \bibinfo{volume}{17},
  \bibinfo{number}{3} (\bibinfo{year}{2021}), \bibinfo{pages}{1--14}.
\newblock


\bibitem[Maaten(2014)]%
        {van2014accelerating}
\bibfield{author}{\bibinfo{person}{Laurens Van~Der Maaten}.}
  \bibinfo{year}{2014}\natexlab{}.
\newblock \showarticletitle{Accelerating t-{SNE} using tree-based algorithms}.
\newblock \bibinfo{journal}{\emph{Journal of Machine Learning Research}}
  (\bibinfo{year}{2014}), \bibinfo{pages}{3221--3245}.
\newblock


\bibitem[Mikolov et~al\mbox{.}(2013)]%
        {mikolov2013distributed}
\bibfield{author}{\bibinfo{person}{Tomas Mikolov}, \bibinfo{person}{Ilya
  Sutskever}, \bibinfo{person}{Kai Chen}, \bibinfo{person}{Greg~S. Corrado},
  {and} \bibinfo{person}{Jeff Dean}.} \bibinfo{year}{2013}\natexlab{}.
\newblock \showarticletitle{Distributed representations of words and phrases
  and their compositionality}. In \bibinfo{booktitle}{\emph{NIPS}}.
  \bibinfo{pages}{3111--3119}.
\newblock


\bibitem[Miller(1995)]%
        {miller1995wordnet}
\bibfield{author}{\bibinfo{person}{George~A. Miller}.}
  \bibinfo{year}{1995}\natexlab{}.
\newblock \showarticletitle{WordNet: a lexical database for English}.
\newblock \bibinfo{journal}{\emph{Commun. ACM}} \bibinfo{volume}{38},
  \bibinfo{number}{11} (\bibinfo{year}{1995}), \bibinfo{pages}{39--41}.
\newblock


\bibitem[Mishra et~al\mbox{.}(2018)]%
        {mishra2018generative}
\bibfield{author}{\bibinfo{person}{Ashish Mishra}, \bibinfo{person}{Shiva
  Krishna~Reddy}, \bibinfo{person}{Anurag Mittal}, {and}
  \bibinfo{person}{Hema~A. Murthy}.} \bibinfo{year}{2018}\natexlab{}.
\newblock \showarticletitle{A generative model for zero shot learning using
  conditional variational autoencoders}. In \bibinfo{booktitle}{\emph{CVPRW}}.
  \bibinfo{pages}{2188--2196}.
\newblock


\bibitem[Narayan et~al\mbox{.}(2020)]%
        {narayan2020latent}
\bibfield{author}{\bibinfo{person}{Sanath Narayan}, \bibinfo{person}{Akshita
  Gupta}, \bibinfo{person}{Fahad~Shahbaz Khan}, \bibinfo{person}{Cees G.~M.
  Snoek}, {and} \bibinfo{person}{Ling Shao}.} \bibinfo{year}{2020}\natexlab{}.
\newblock \showarticletitle{Latent embedding feedback and discriminative
  features for zero-shot classification}. In \bibinfo{booktitle}{\emph{ECCV}}.
  \bibinfo{pages}{479--495}.
\newblock


\bibitem[Norouzi et~al\mbox{.}(2013)]%
        {norouzi2013zero}
\bibfield{author}{\bibinfo{person}{Mohammad Norouzi},
  \bibinfo{person}{Tom{\'{a}}s Mikolov}, \bibinfo{person}{Samy Bengio},
  \bibinfo{person}{Yoram Singer}, \bibinfo{person}{Jonathon Shlens},
  \bibinfo{person}{Andrea Frome}, \bibinfo{person}{Greg Corrado}, {and}
  \bibinfo{person}{Jeffrey Dean}.} \bibinfo{year}{2013}\natexlab{}.
\newblock \showarticletitle{Zero-shot learning by convex combination of
  semantic embeddings}.
\newblock \bibinfo{journal}{\emph{arXiv preprint arXiv:1312.5650}}
  (\bibinfo{year}{2013}).
\newblock


\bibitem[Patterson et~al\mbox{.}(2014)]%
        {patterson2014sun}
\bibfield{author}{\bibinfo{person}{Genevieve Patterson}, \bibinfo{person}{Chen
  Xu}, \bibinfo{person}{Hang Su}, {and} \bibinfo{person}{James Hays}.}
  \bibinfo{year}{2014}\natexlab{}.
\newblock \showarticletitle{The {SUN} Attribute Database: Beyond Categories for
  Deeper Scene Understanding}.
\newblock \bibinfo{journal}{\emph{IJCV}} \bibinfo{volume}{108},
  \bibinfo{number}{1-2} (\bibinfo{year}{2014}), \bibinfo{pages}{59–--81}.
\newblock


\bibitem[Pennington et~al\mbox{.}(2014)]%
        {pennington-etal-2014-glove}
\bibfield{author}{\bibinfo{person}{Jeffrey Pennington},
  \bibinfo{person}{Richard Socher}, {and} \bibinfo{person}{Christopher
  Manning}.} \bibinfo{year}{2014}\natexlab{}.
\newblock \showarticletitle{{G}lo{V}e: Global Vectors for Word Representation}.
  In \bibinfo{booktitle}{\emph{Proceedings of the 2014 Conference on Empirical
  Methods in Natural Language Processing (EMNLP)}}.
  \bibinfo{pages}{1532--1543}.
\newblock


\bibitem[Rahman et~al\mbox{.}(2021)]%
        {rahman2021multimodal}
\bibfield{author}{\bibinfo{person}{Md~Abdur Rahman}, \bibinfo{person}{M~Shamim
  Hossain}, \bibinfo{person}{Nabil~A Alrajeh}, {and} \bibinfo{person}{BB
  Gupta}.} \bibinfo{year}{2021}\natexlab{}.
\newblock \showarticletitle{A multimodal, multimedia point-of-care deep
  learning framework for COVID-19 diagnosis}.
\newblock \bibinfo{journal}{\emph{ACM Transactions on Multimidia Computing
  Communications and Applications}} \bibinfo{volume}{17}, \bibinfo{number}{1s}
  (\bibinfo{year}{2021}), \bibinfo{pages}{1--24}.
\newblock


\bibitem[Rajasekhar and Jaswal(2015)]%
        {rajasekhar2015autonomous}
\bibfield{author}{\bibinfo{person}{MV Rajasekhar} {and}
  \bibinfo{person}{Anil~Kumar Jaswal}.} \bibinfo{year}{2015}\natexlab{}.
\newblock \showarticletitle{Autonomous vehicles: The future of automobiles}. In
  \bibinfo{booktitle}{\emph{2015 IEEE International Transportation
  Electrification Conference (ITEC)}}. IEEE, \bibinfo{pages}{1--6}.
\newblock


\bibitem[Rezaei and Klette(2014)]%
        {rezaei2014look}
\bibfield{author}{\bibinfo{person}{Mahdi Rezaei} {and}
  \bibinfo{person}{Reinhard Klette}.} \bibinfo{year}{2014}\natexlab{}.
\newblock \showarticletitle{Look at the driver, look at the road: No
  distraction! no accident!}. In \bibinfo{booktitle}{\emph{Proceedings of the
  IEEE conference on computer vision and pattern recognition}}.
  \bibinfo{pages}{129--136}.
\newblock


\bibitem[Rezaei and Shahidi(2020)]%
        {rezaei2020zero}
\bibfield{author}{\bibinfo{person}{Mahdi Rezaei} {and} \bibinfo{person}{Mahsa
  Shahidi}.} \bibinfo{year}{2020}\natexlab{}.
\newblock \showarticletitle{Zero-shot learning and its applications from
  autonomous vehicles to {COVID}-19 diagnosis: {A} review}.
\newblock \bibinfo{journal}{\emph{Intelligence-based medicine}}
  (\bibinfo{year}{2020}), \bibinfo{pages}{100005}.
\newblock


\bibitem[Rohrbach et~al\mbox{.}(2011)]%
        {rohrbach2011evaluating}
\bibfield{author}{\bibinfo{person}{Marcus Rohrbach}, \bibinfo{person}{Michael
  Stark}, {and} \bibinfo{person}{Bernt Schiele}.}
  \bibinfo{year}{2011}\natexlab{}.
\newblock \showarticletitle{Evaluating knowledge transfer and zero-shot
  learning in a large-scale setting}. In \bibinfo{booktitle}{\emph{CVPR}}.
  \bibinfo{pages}{1641--1648}.
\newblock


\bibitem[Rohrbach et~al\mbox{.}(2010)]%
        {rohrbach2010helps}
\bibfield{author}{\bibinfo{person}{Marcus Rohrbach}, \bibinfo{person}{Michael
  Stark}, \bibinfo{person}{Gy{\"o}rgy Szarvas}, \bibinfo{person}{Iryna
  Gurevych}, {and} \bibinfo{person}{Bernt Schiele}.}
  \bibinfo{year}{2010}\natexlab{}.
\newblock \showarticletitle{What helps where--and why? semantic relatedness for
  knowledge transfer}. In \bibinfo{booktitle}{\emph{CVPR}}.
  \bibinfo{pages}{910--917}.
\newblock


\bibitem[Romera-Paredes and Torr(2015)]%
        {romera2015embarrassingly}
\bibfield{author}{\bibinfo{person}{Bernardino Romera-Paredes} {and}
  \bibinfo{person}{Philip H.~S. Torr}.} \bibinfo{year}{2015}\natexlab{}.
\newblock \showarticletitle{An embarrassingly simple approach to zero-shot
  learning}. In \bibinfo{booktitle}{\emph{Proceedings of the 32nd International
  Conference on International Conference on Machine Learning}}.
  \bibinfo{pages}{2152--2161}.
\newblock


\bibitem[Rousseeuw(1987)]%
        {ROUSSEEUW198753}
\bibfield{author}{\bibinfo{person}{Peter~J. Rousseeuw}.}
  \bibinfo{year}{1987}\natexlab{}.
\newblock \showarticletitle{Silhouettes: a graphical aid to the interpretation
  and validation of cluster analysis}.
\newblock \bibinfo{journal}{\emph{Journal of computational and applied
  mathematics}}  \bibinfo{volume}{20} (\bibinfo{year}{1987}),
  \bibinfo{pages}{53--65}.
\newblock


\bibitem[Russakovsky et~al\mbox{.}(2015)]%
        {russakovsky2015imagenet}
\bibfield{author}{\bibinfo{person}{Olga Russakovsky}, \bibinfo{person}{Jia
  Deng}, \bibinfo{person}{Hao Su}, \bibinfo{person}{Jonathan Krause},
  \bibinfo{person}{Sanjeev Satheesh}, \bibinfo{person}{Sean Ma},
  \bibinfo{person}{Zhiheng Huang}, \bibinfo{person}{Andrej Karpathy},
  \bibinfo{person}{Aditya Khosla}, \bibinfo{person}{Michael Bernstein},
  {et~al\mbox{.}}} \bibinfo{year}{2015}\natexlab{}.
\newblock \showarticletitle{Image{N}et {L}arge {S}cale {V}isual {R}ecognition
  {C}hallenge}.
\newblock \bibinfo{journal}{\emph{IJCV}} \bibinfo{volume}{115},
  \bibinfo{number}{3} (\bibinfo{year}{2015}), \bibinfo{pages}{211--252}.
\newblock


\bibitem[Simonyan and Zisserman(2014)]%
        {simonyan2014very}
\bibfield{author}{\bibinfo{person}{Karen Simonyan} {and}
  \bibinfo{person}{Andrew Zisserman}.} \bibinfo{year}{2014}\natexlab{}.
\newblock \showarticletitle{Very deep convolutional networks for large-scale
  image recognition}.
\newblock \bibinfo{journal}{\emph{arXiv preprint arXiv:1409.1556}}
  (\bibinfo{year}{2014}).
\newblock


\bibitem[Socher et~al\mbox{.}(2013)]%
        {socher2013zero}
\bibfield{author}{\bibinfo{person}{Richard Socher}, \bibinfo{person}{Milind
  Ganjoo}, \bibinfo{person}{Christopher~D. Manning}, {and}
  \bibinfo{person}{Andrew Ng}.} \bibinfo{year}{2013}\natexlab{}.
\newblock \showarticletitle{Zero-shot learning through cross-modal transfer}.
  In \bibinfo{booktitle}{\emph{NIPS}}. \bibinfo{pages}{935--943}.
\newblock


\bibitem[Szegedy et~al\mbox{.}(2015)]%
        {szegedy2015going}
\bibfield{author}{\bibinfo{person}{Christian Szegedy}, \bibinfo{person}{Wei
  Liu}, \bibinfo{person}{Yangqing Jia}, \bibinfo{person}{Pierre Sermanet},
  \bibinfo{person}{Scott Reed}, \bibinfo{person}{Dragomir Anguelov},
  \bibinfo{person}{Dumitru Erhan}, \bibinfo{person}{Vincent Vanhoucke}, {and}
  \bibinfo{person}{Andrew Rabinovich}.} \bibinfo{year}{2015}\natexlab{}.
\newblock \showarticletitle{Going deeper with convolutions}. In
  \bibinfo{booktitle}{\emph{CVPR}}. \bibinfo{pages}{1--9}.
\newblock


\bibitem[Tang et~al\mbox{.}(2021)]%
        {tang2021zero}
\bibfield{author}{\bibinfo{person}{Chenwei Tang}, \bibinfo{person}{Zhenan He},
  \bibinfo{person}{Yunxia Li}, {and} \bibinfo{person}{Jiancheng Lv}.}
  \bibinfo{year}{2021}\natexlab{}.
\newblock \showarticletitle{Zero-Shot Learning via Structure-Aligned Generative
  Adversarial Network}.
\newblock \bibinfo{journal}{\emph{{IEEE} Transactions on Neural Networks and
  Learning Systems}} (\bibinfo{year}{2021}), \bibinfo{pages}{1--14}.
\newblock


\bibitem[Tomanek et~al\mbox{.}(2009)]%
        {tomanek-etal-2009-proper}
\bibfield{author}{\bibinfo{person}{Katrin Tomanek}, \bibinfo{person}{Florian
  Laws}, \bibinfo{person}{Udo Hahn}, {and} \bibinfo{person}{Hinrich
  Sch{\"u}tze}.} \bibinfo{year}{2009}\natexlab{}.
\newblock \showarticletitle{On proper unit selection in active learning:
  co-selection effects for named entity recognition}. In
  \bibinfo{booktitle}{\emph{Proceedings of the {NAACL} {HLT} Workshop on Active
  Learning for Natural Language Processing}}. \bibinfo{pages}{9--17}.
\newblock


\bibitem[Vyas et~al\mbox{.}(2020)]%
        {vyas2020leveraging}
\bibfield{author}{\bibinfo{person}{Maunil~R. Vyas}, \bibinfo{person}{Hemanth
  Venkateswara}, {and} \bibinfo{person}{Sethuraman Panchanathan}.}
  \bibinfo{year}{2020}\natexlab{}.
\newblock \showarticletitle{Leveraging seen and unseen semantic relationships
  for generative zero-shot learning}. In \bibinfo{booktitle}{\emph{ECCV}}.
  \bibinfo{pages}{70--86}.
\newblock


\bibitem[Wah. et~al\mbox{.}(2011)]%
        {WahCUB_200_2011}
\bibfield{author}{\bibinfo{person}{C. Wah.}, \bibinfo{person}{S. Branson},
  \bibinfo{person}{P. Welinder}, \bibinfo{person}{P. Perona}, {and}
  \bibinfo{person}{S. Belongie}.} \bibinfo{year}{2011}\natexlab{}.
\newblock \bibinfo{booktitle}{\emph{The {Caltech-UCSD Birds}-200-2011
  Dataset}}.
\newblock \bibinfo{type}{{T}echnical {R}eport} CNS-TR-2011-001.
  \bibinfo{institution}{California Institute of Technology}.
\newblock


\bibitem[Wang et~al\mbox{.}(2021)]%
        {wang2021graph}
\bibfield{author}{\bibinfo{person}{Qunbo Wang}, \bibinfo{person}{Wenjun Wu},
  \bibinfo{person}{Yongchi Zhao}, {and} \bibinfo{person}{Yuzhang Zhuang}.}
  \bibinfo{year}{2021}\natexlab{}.
\newblock \showarticletitle{Graph active learning for GCN-based zero-shot
  classification}.
\newblock \bibinfo{journal}{\emph{Neurocomputing}}  \bibinfo{volume}{435}
  (\bibinfo{year}{2021}), \bibinfo{pages}{15--25}.
\newblock


\bibitem[Ward(1963)]%
        {ward1963hierarchical}
\bibfield{author}{\bibinfo{person}{Joe~H. Ward, Jr.}}
  \bibinfo{year}{1963}\natexlab{}.
\newblock \showarticletitle{Hierarchical Grouping to Optimize an Objective
  Function}.
\newblock \bibinfo{journal}{\emph{Journal of the American statistical
  association}} \bibinfo{volume}{58}, \bibinfo{number}{301}
  (\bibinfo{year}{1963}), \bibinfo{pages}{236--244}.
\newblock


\bibitem[Xian et~al\mbox{.}(2016)]%
        {xian2016latent}
\bibfield{author}{\bibinfo{person}{Yongqin Xian}, \bibinfo{person}{Zeynep
  Akata}, \bibinfo{person}{Gaurav Sharma}, \bibinfo{person}{Quynh Nguyen},
  \bibinfo{person}{Matthias Hein}, {and} \bibinfo{person}{Bernt Schiele}.}
  \bibinfo{year}{2016}\natexlab{}.
\newblock \showarticletitle{Latent embeddings for zero-shot classification}. In
  \bibinfo{booktitle}{\emph{CVPR}}. \bibinfo{pages}{69--77}.
\newblock


\bibitem[Xian et~al\mbox{.}(2018a)]%
        {xian2018zero}
\bibfield{author}{\bibinfo{person}{Yongqin Xian}, \bibinfo{person}{Christoph~H
  Lampert}, \bibinfo{person}{Bernt Schiele}, {and} \bibinfo{person}{Zeynep
  Akata}.} \bibinfo{year}{2018}\natexlab{a}.
\newblock \showarticletitle{Zero-shot learning—A comprehensive evaluation of
  the good, the bad and the ugly}.
\newblock \bibinfo{journal}{\emph{PAMI}} \bibinfo{volume}{41},
  \bibinfo{number}{9} (\bibinfo{year}{2018}), \bibinfo{pages}{2251--2265}.
\newblock


\bibitem[Xian et~al\mbox{.}(2018b)]%
        {xian2018feature}
\bibfield{author}{\bibinfo{person}{Yongqin Xian}, \bibinfo{person}{Tobias
  Lorenz}, \bibinfo{person}{Bernt Schiele}, {and} \bibinfo{person}{Zeynep
  Akata}.} \bibinfo{year}{2018}\natexlab{b}.
\newblock \showarticletitle{Feature generating networks for zero-shot
  learning}. In \bibinfo{booktitle}{\emph{CVPR}}. \bibinfo{pages}{5542--5551}.
\newblock


\bibitem[Xian et~al\mbox{.}(2019)]%
        {xian2019f}
\bibfield{author}{\bibinfo{person}{Yongqin Xian}, \bibinfo{person}{Saurabh
  Sharma}, \bibinfo{person}{Bernt Schiele}, {and} \bibinfo{person}{Zeynep
  Akata}.} \bibinfo{year}{2019}\natexlab{}.
\newblock \showarticletitle{{F-VAEGAN-D2}: {A} feature generating framework for
  any-shot learning}. In \bibinfo{booktitle}{\emph{CVPR}}.
  \bibinfo{pages}{10267--10276}.
\newblock


\bibitem[Xie and Philip(2017)]%
        {xie2017active}
\bibfield{author}{\bibinfo{person}{Sihong Xie} {and} \bibinfo{person}{S~Yu
  Philip}.} \bibinfo{year}{2017}\natexlab{}.
\newblock \showarticletitle{Active zero-shot learning: a novel approach to
  extreme multi-labeled classification}.
\newblock \bibinfo{journal}{\emph{International Journal of Data Science and
  Analytics}} \bibinfo{volume}{3}, \bibinfo{number}{3} (\bibinfo{year}{2017}),
  \bibinfo{pages}{151--160}.
\newblock


\bibitem[Xie et~al\mbox{.}(2016)]%
        {xie2016active}
\bibfield{author}{\bibinfo{person}{Sihong Xie}, \bibinfo{person}{Shaoxiong
  Wang}, {and} \bibinfo{person}{Philip~S Yu}.} \bibinfo{year}{2016}\natexlab{}.
\newblock \showarticletitle{Active zero-shot learning}. In
  \bibinfo{booktitle}{\emph{Proceedings of the 25th ACM International on
  Conference on Information and Knowledge Management}}.
  \bibinfo{pages}{1889--1892}.
\newblock


\bibitem[Xu et~al\mbox{.}(2021)]%
        {xu2021zero}
\bibfield{author}{\bibinfo{person}{Xing Xu}, \bibinfo{person}{Jialin Tian},
  \bibinfo{person}{Kaiyi Lin}, \bibinfo{person}{Huimin Lu},
  \bibinfo{person}{Jie Shao}, {and} \bibinfo{person}{Heng~Tao Shen}.}
  \bibinfo{year}{2021}\natexlab{}.
\newblock \showarticletitle{Zero-shot cross-modal retrieval by assembling
  AutoEncoder and generative adversarial network}.
\newblock \bibinfo{journal}{\emph{ACM Transactions on Multimedia Computing,
  Communications, and Applications (TOMM)}} \bibinfo{volume}{17},
  \bibinfo{number}{1s} (\bibinfo{year}{2021}), \bibinfo{pages}{1--17}.
\newblock


\bibitem[Zhang and Saligrama(2015)]%
        {zhang2015zero}
\bibfield{author}{\bibinfo{person}{Ziming Zhang} {and}
  \bibinfo{person}{Venkatesh Saligrama}.} \bibinfo{year}{2015}\natexlab{}.
\newblock \showarticletitle{Zero-shot learning via semantic similarity
  embedding}. In \bibinfo{booktitle}{\emph{ICCV}}. \bibinfo{pages}{4166--4174}.
\newblock


\end{thebibliography}

%
%
%
%
%
%
%
%

\end{document}